\documentclass[sigconf, screen=true, review=false, printacmref=false, printccs=false, printfolios=false]{acmart}
\usepackage{comment}

\usepackage{booktabs} 
\usepackage{multibib}
\usepackage{xcolor}
\newcites{latex}{Additional References}
\usepackage{mdwlist}
\setcopyright{none}

\usepackage{hyperref}
\hypersetup{
colorlinks=true,
linkcolor=black,
citecolor=black,
filecolor=mydarkblue,
urlcolor=mydarkblue}

\usepackage{enumitem}

\newtheorem{property}{Property}

\newtheorem{Definition}{Definition}
\newtheorem{Claim}{Claim}

\newtheorem{Result}{\textbf{\textsc{Result}}}

\usepackage{footmisc}
\usepackage{comment}
\usepackage{tabularx}
\usepackage{mathrsfs}
\usepackage{amsmath}
\usepackage{algorithm}
\usepackage{footmisc}
\usepackage{tikz}
\usepackage{xspace}
\usepackage{graphicx}
\usepackage{multirow}
\usepackage{subcaption}
\usepackage{pifont}
\newcommand{\cmark}{\ding{51}}
\newcommand{\xmark}{\ding{55}}
\usetikzlibrary{matrix,decorations.pathreplacing,calc}
\usepackage{mathtools}

\pgfkeys{tikz/mymatrixenv/.style={decoration=brace,every left delimiter/.style={xshift=3pt},every right delimiter/.style={xshift=-3pt}}}
\pgfkeys{tikz/mymatrix/.style={matrix of math nodes,left delimiter=[,right delimiter={]},inner sep=2pt,column sep=1em,row sep=0.5em,nodes={inner sep=0pt}}}
\pgfkeys{tikz/mymatrixbrace/.style={decorate,thick}}

\usepackage{xcolor}
\definecolor{thedarkblue}{RGB}{0,0,120} 
\definecolor{mydarkblue}{rgb}{0,0.08,0.45}

\usepackage{microtype}
\usepackage{balance}

\newcolumntype{L}[1]{>{\raggedright\let\newline\\\arraybackslash\hspace{0pt}}m{#1}}
\newcolumntype{C}[1]{>{\centering\let\newline\\\arraybackslash\hspace{0pt}}m{#1}}
\newcolumntype{R}[1]{>{\raggedleft\let\newline\\\arraybackslash\hspace{0pt}}m{#1}}

\providecommand{\mat}[1]{\boldsymbol{\mathrm{#1}}}
\renewcommand{\vec}[1]{\boldsymbol{\mathrm{#1}}}

\DeclareMathOperator{\hugeE}{\mbox{\huge\raise-0.3ex\hbox{E}}}
\DeclareMathOperator{\p}{\mathbb{P}}
\DeclareMathOperator{\hugep}{\mbox{\huge\raise-0.3ex\hbox{$\p$}}}

\newcommand{\RR}{\mathbb{R}}

\newcommand{\methodweakT}{\textsc{WTRG-$\tau$}\xspace}
\newcommand{\methodweakN}{\textsc{WTRG-$\epsilon$}\xspace}

\providecommand{\mA}{\ensuremath{\mat{A}}}

\providecommand{\mS}{\ensuremath{\mat{S}}}

\providecommand{\mZ}{\ensuremath{\mat{Z}}}

\providecommand{\vy}{\ensuremath{\vec{y}}}
\providecommand{\vz}{\ensuremath{\vec{z}}}

\usepackage{setspace}
\graphicspath{{./}{./graphics/}}
\newcolumntype{H}{>{\setbox0=\hbox\bgroup}c<{\egroup}@{}}

\newcommand{\eg}{\emph{e.g.}}
\newcommand{\ie}{\emph{i.e.}}

\newcommand\TT{\rule{0pt}{2.7ex}}

\usepackage{tikz}
\usepackage{verbatim}
\usetikzlibrary{arrows}
\usetikzlibrary{shapes,snakes}
\usetikzlibrary{decorations.pathmorphing} 
\usetikzlibrary{fit}					
\usetikzlibrary{backgrounds}	

\usepackage{algorithm}
\usepackage{algorithmicx}
\usepackage{algpseudocode}

\algblockdefx[parallel]{ParFor}{EndPar}[1][]{$\textbf{parallel for}$ #1 $\textbf{do}$}{$\textbf{end parallel}$}
\algrenewcommand{\alglinenumber}[1]{\fontsize{6.5}{7}\selectfont#1}
\algtext*{EndPar}

\algblockdefx[parallel]{parfor}{endpar}[1][]{$\textbf{parallel for}$ #1 $\textbf{do}$}{$\textbf{end parallel}$}
\algrenewcommand{\alglinenumber}[1]{\scriptsize#1:}

\algblockdefx[foreach1]{foreach}{endforeach}[1][]{$\textbf{for}$ #1 $\textbf{do}$}{$\textbf{end for}$}
\algrenewcommand{\alglinenumber}[1]{\scriptsize#1:}

\algnotext{EndFor}
\algnotext{EndIf}
\algnotext{EndProcedure}



\algblockdefx[fornew1]{foreach}{endforeach}
[1][]{\textbf{for} #1}
{\textbf{end for}}

\newcommand{\multiline}[1]{\State \parbox[t]{\dimexpr\linewidth-\algorithmicindent}{#1\strut}}

\begin{document}

\title{From Static to Dynamic Node Embeddings}

\settopmatter{authorsperrow=4}
\author{Di Jin}
\affiliation{
\institution{University of Michigan}
}
\email{dijin@umich.edu}

\author{Sungchul Kim}
\affiliation{
\institution{Adobe Research}
}
\email{sukim@adobe.com}
\email{}

\author{Ryan A. Rossi}
\orcid{1234-5678-9012-3456}
\affiliation{
\institution{Adobe Research}
}
\email{ryrossi@adobe.com}

\author{Danai Koutra}
\affiliation{
\institution{University of Michigan}
}
\email{dkoutra@umich.edu}
\email{}

\renewcommand{\shortauthors}{D.~Jin, S.~Kim, R.~A.~Rossi, and D. Koutra}

\begin{abstract}
We introduce a general framework for leveraging graph stream data for temporal prediction-based applications. Our proposed framework includes novel methods for learning an appropriate graph time-series representation, modeling and weighting the temporal dependencies, and generalizing existing embedding methods for such data. While previous work on dynamic modeling and embedding have focused on representing a stream of timestamped edges using a time-series of graphs based on a specific time-scale (\eg, 1 month), we propose the notion of an $\epsilon$-graph time-series that uses a fixed number of edges for each graph, and show its superiority over the time-scale representation used in previous work. In addition, we propose a number of new temporal models based on the notion of temporal reachability graphs and weighted temporal summary graphs. These temporal models are then used to generalize existing base (static) embedding methods by enabling them to incorporate and appropriately model temporal dependencies in the data. From the 6 temporal network models investigated (for each of the 7 base embedding methods), we find that the top-3 temporal models are always those that leverage the new $\epsilon$-graph time-series representation. Furthermore, the dynamic embedding methods from the framework almost always achieve better predictive performance than existing state-of-the-art dynamic node embedding methods that are developed specifically for such temporal prediction tasks. Finally, the findings of this work are useful for designing better dynamic embedding methods.
\end{abstract}


\maketitle

\section{Introduction}
\label{sec:intro}
Real-world networks that record the interaction between entities have grown rapidly, for example, the Internet~\cite{coffman2002growth,chang2003internet-o42}, various online social networks (\eg, Facebook, Snapchat), citation and collaboration networks in academia~\cite{leskovec2005graphs}. 
Specifically, when nodes and edges continuously change over time with addition, deletion (e.g., a phone call, an email, or physical proximity between two entities at a specific point in time), we have a particular type of evolving network structure.
Learning an appropriate network representation (embedding) that accurately captures the temporal dynamics and temporal structural properties of these entities is important for many downstream machine learning tasks such as recommendation, entity resolution, link prediction, among many others.

\begin{figure}[t!]
\vspace{1mm}
\centering
\includegraphics[width=0.49\linewidth]{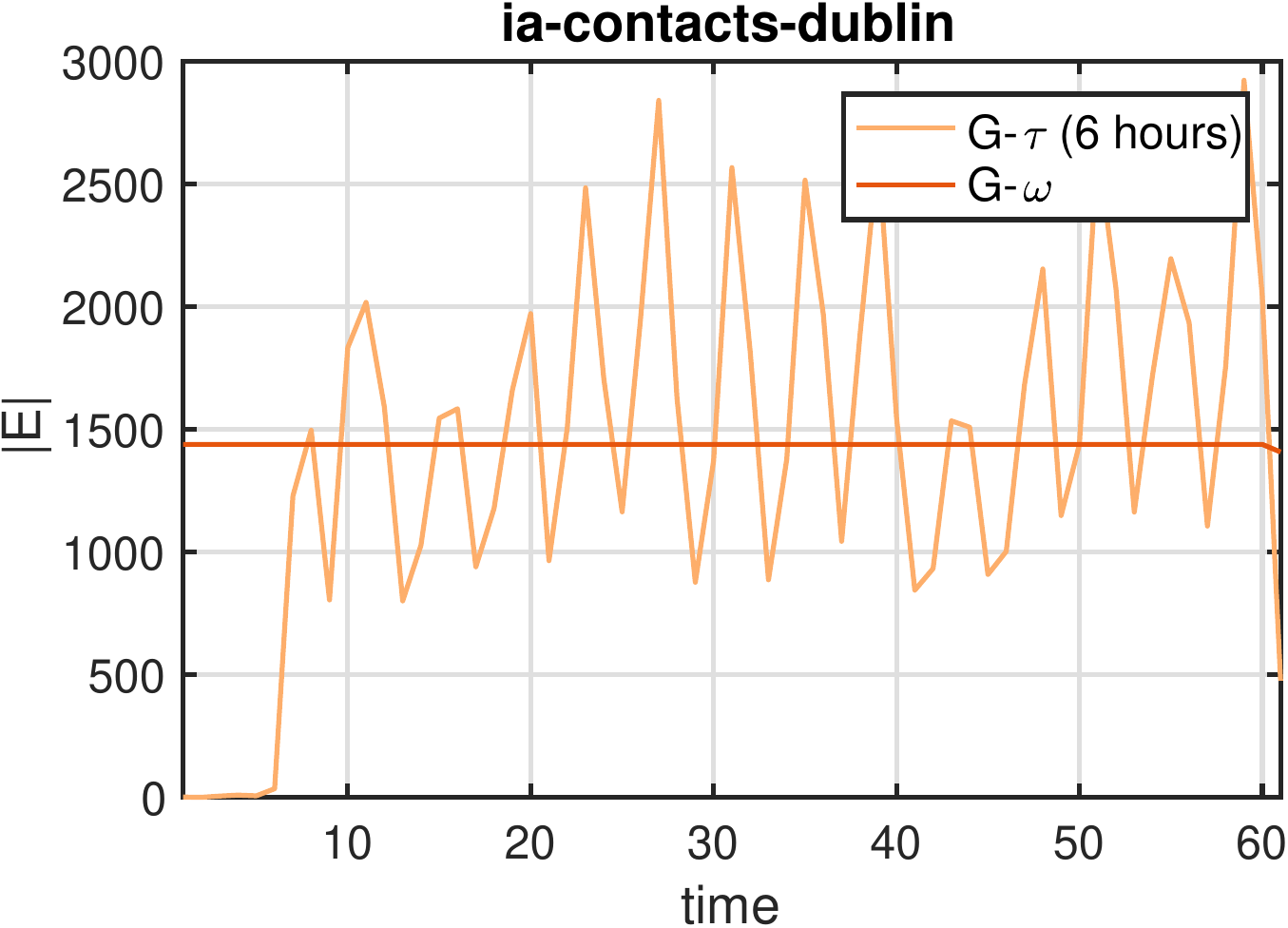}\hfill
\includegraphics[width=0.485\linewidth]{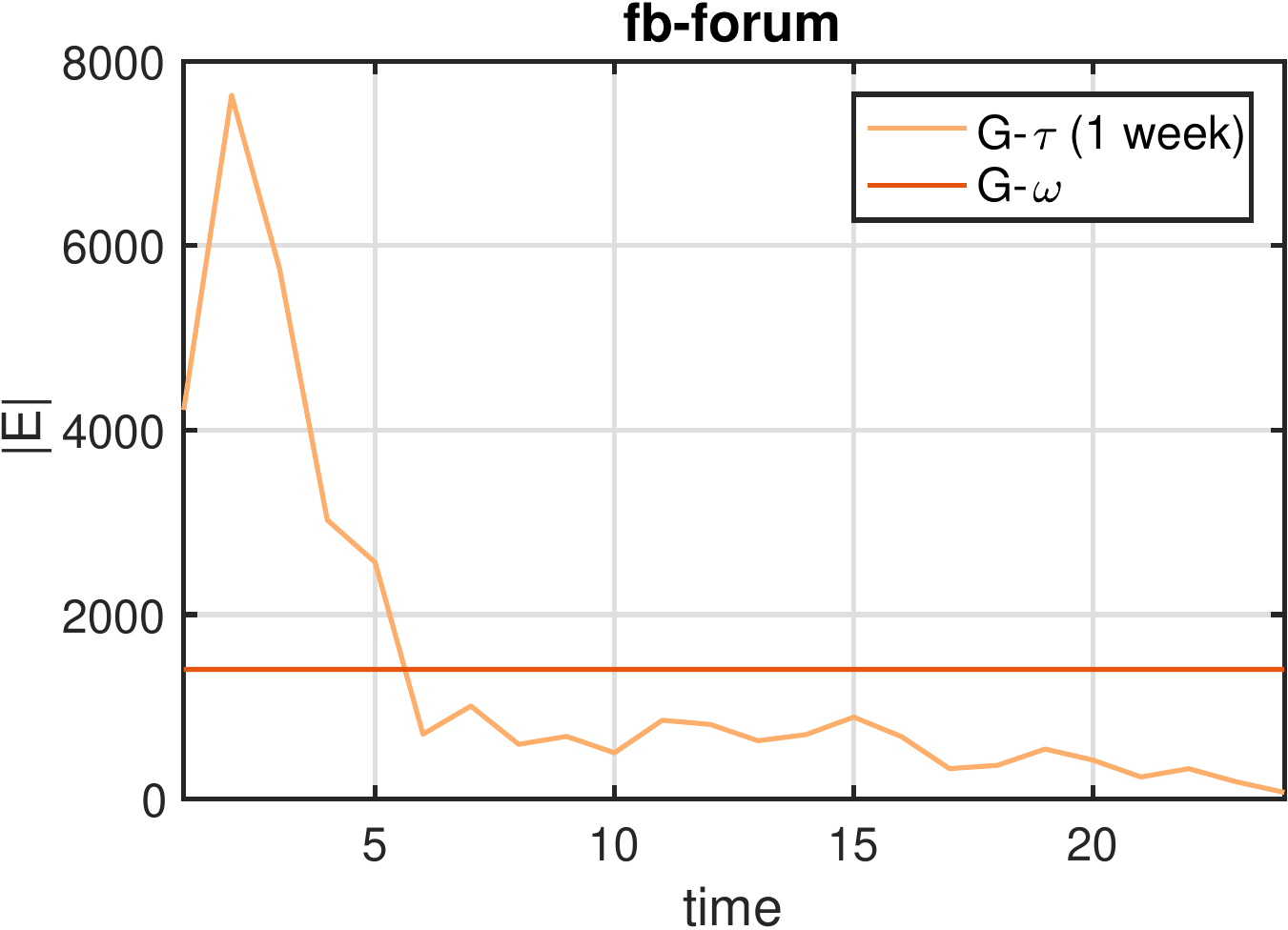}
\vspace{-2mm}
\caption{Comparing the $\tau$-graph time-series (where each graph contains edges within a time-scale, \ie, 6 hrs and 1 week, respectively) and the $\epsilon$-graph time-series representation (where each graph contains the fixed \#edges) with respect to the number of temporal edges over time.
}
\label{fig:properties-comparing-graph-time-series}
\vspace{-5mm}
\end{figure}

Recently, significant research efforts have been devoted in the field of temporal representation learning, most of which follow the same pipeline: given a time-series of graphs, $\mathcal{G}=\{G_1, G_2, \cdots, G_T\}$, modeling the individual graph structures (intra-snapshot property) along with the temporal dependency (inter-snapshot relation), and deriving node embeddings that incorporate both perspectives.
It is worth noting that these works achieve promising performance at the cost of time and model complexity, which limits their usage on large temporal graphs. For example, introducing extra transition variables to reflect the temporal dependency between snapshots~\cite{goyal2018dyngem}, or latent weights on edges between snapshots~\cite{sankar2020dysat}.

In this work, we propose a general framework that can be used as a black-box to generalize any static embedding method to a more powerful and predictive dynamic embedding method.
Empirically, we observe that our framework achieves similar or even higher performance on prediction tasks, with much less time and model complexity.
The framework consists of three main components: 
(\textbf{C1}) a graph time-series representation, 
(\textbf{C2}) a temporal network model that appropriately models and weights the temporal dependencies in the graph time-series, and
(\textbf{C3}) a base embedding method to learn a time-series of embeddings along with a temporal fusion mechanism to derive the final temporal node embeddings.
The framework is highly expressive as any unique combination of $\mathbf{C1}$-$\mathbf{C3}$ gives rise to a new dynamic embedding method.

While previous works on dynamic modeling and embedding have focused on representing the stream of timestamped edges~\cite{CTDNE-WWW18} using a time-series of graphs based on a specific time-scale $\tau$ (\eg, $\tau=1$ hour, or $1$ month)~\cite{goyal2018dyngem,goyal2019dyngraph2vec,ijcai2019-640,leskovec2005graphs,zhou2018dynamic,sankar2020dysat}, we instead propose the notion of an $\epsilon$-graph time-series that uses a fixed number of edges for each graph in the time-series.
Theoretically, by fixing the number of edges to be $\epsilon$ in each graph, we ensure that every graph in the sequence has an equal probability of giving rise to the same distribution of higher-order network motifs (graphlets)~\cite{ahmed2019temporal} and other structural patterns\footnote{This is in contrast to graphs with different amounts of edges.  As an example, suppose we have two arbitrary graphs $G_{t-1}=(V_{t-1},E_{t-1})$ and $G_t=(V_{t},E_{t})$ where $|E_{t-1}| \ll |E_{t}|$, then from the very beginning, we know that the counts of all $k \in \{3,4,\ldots\}$-node network motifs (graphlets) in $G_t$ is almost surely larger than $G_{t-1}$.}, and therefore, the new $\epsilon$-graph time-series forces the models to avoid capturing simple trivial differences due to edge counts, and instead, allow the models to capture actual \emph{structural changes} to the graphs over time.
In other words, since the proposed $\epsilon$-graph time-series controls for the number of edges over time, embedding methods can more appropriately model and capture the actual change in the structural properties and subgraph patterns over time, as opposed to just the frequency of edges that is captured by the $\tau$-graph time-series representation used in previous work. 
Another advantage of the $\epsilon$-graph time-series representation is that it preserves the sequential order of timestamped edges \textit{without} suffering from the structural instability of the graph due to the sometimes drastic difference in edge counts from one time to the next.
As observed in Fig.~\ref{fig:properties-comparing-graph-time-series}, while the $\epsilon$-graph time-series representation has a fixed number of edges over time, conventionally-used $\tau$ representation can significantly deviate with large spikes even between consecutive graphs in the series.
Finally, it is obvious that if a graph time-series representation is unable to capture the simplest 1st-order subgraph structures (edges), then by definition it cannot capture higher-order subgraph structures that are built on such lower-order ones.
Hence, the proposed $\epsilon$-graph time-series representation should be used if the goal is to model the \emph{structural changes} between graphs whereas the $\tau$-graph time-series should be used if the goal is to capture changes in \emph{edge frequencies} for a fixed application-specific time-scale such as 10 minutes or 1 hour.

We also introduce a number of important temporal models that can be leveraged over any graph time-series representation of the edge stream.
The first temporal model is based on the notion of a temporal reachability graph (TRG).
A temporal reachability graph (TRG) is derived by transforming a dynamic graph into a static graph where an edge from $u$ to $w$ indicates a temporal walk. The second temporal model that we introduce is called a weighted temporal summary graph. Notably, a weighted temporal summary graph captures the temporal recurrence and temporal recency of links by appropriately weighting links with respect to a function $f$ that assigns larger weights to links that are more recent and recurrent whereas links that occur in the more distant past are assigned lower weights. All temporal models can leverage either $\{\epsilon,\tau\}$-graph time-series representation.

This paper aims to provide a systematic exploration of the most useful graph time-series representations and temporal network models (used to incorporate the temporal dependencies into base embedding methods) in downstream temporal prediction tasks.
To the best of our knowledge, this is the first work of this kind. 
The proposed framework provides a basis for investigating different graph time-series representation as well as temporal network models.
Furthermore, it can also be used as a basis for generalizing static embedding methods for handling and modeling temporal dependencies.
The main contributions of this work are as follows:
\begin{itemize}
\item \textbf{General Framework:} 
We describe a general framework for leveraging graph stream data for temporal prediction-based applications that can generalize any static graph embedding method.
The proposed framework includes novel methods for learning an appropriate graph time-series representation, modeling and weighting the temporal dependencies in such data, and generalizing existing embedding methods for such data.
The framework can be used as a blackbox to obtain new dynamic node embedding methods that are significantly better and more accurate for such massive streaming graph data.

\item \textbf{Novel Graph Time-series Representation:} 
We introduce the notion of a $\epsilon$-graph time-series where each graph in the time-series has a fixed $\epsilon$ number of edges.
We show that the proposed $\epsilon$-graph time-series representation has a number of useful properties compared to the conventional way of discretizing the edge stream based on the application time-scale such as 1 hour, 1 day, or 1 month.

\item \textbf{New Temporal Network Models:} 
We propose a number of new temporal network models and use them to extend $7$ existing static embedding methods.
Furthermore, we find that the top-3 temporal models (across all graphs and base embedding methods investigated) are those that leverage the proposed $\epsilon$-graph time-series representation.

\item \textbf{Comprehensive Evaluation:}
Strikingly, the dynamic embedding methods from the general framework achieve better predictive performance than existing state-of-the-art dynamic node embedding methods that are significantly more complex and developed specifically for such temporal prediction tasks.
These results demonstrate the utility of the proposed framework and motivates its use in future research for developing better dynamic node embedding methods as well as evaluating the utility of more sophisticated and complex methods.
\end{itemize}

\begin{table}[b]
\centering
\caption{Qualitative comparison of existing embedding methods on temporal graphs. 
The graph time-series representation used by the method (application time-scale, or fixed number of edges), 
the type of temporal model used, 
and type of embedding fusion used (if any).
}
\label{table_comp}
\vspace{-0.2cm}
\resizebox{\columnwidth}{!}{
\footnotesize
\setlength{\tabcolsep}{2.6pt} 
\begin{tabularx}{1.0\linewidth}{@{}l@{} cc cc cc c HH @{}}
\toprule
& 
\multicolumn{2}{c}{\textsc{Representation}} & 
\multicolumn{2}{c}{\textsc{Temporal Model}} &
\\

\cmidrule(lr){2-3}
\cmidrule(lr){4-5}

& 
Time ($\tau$) & 
Edge ($\epsilon$) & 
Snapshot &
Weighting & 
\textsc{Embedding Fusion} &
\\

\midrule
DANE~\cite{li2017attributed} & \cmark & \xmark & \cmark & \xmark  & \xmark \\
DynGem~\cite{goyal2018dyngem} & \cmark & \xmark & \cmark & \xmark  & \cmark \\
TIMERS~\cite{zhang2018timers} & \cmark & \xmark & \cmark  & \xmark & \xmark \\
Dynagraph2vec~\cite{goyal2019dyngraph2vec} & \cmark & \xmark &  \cmark & \xmark  & \cmark \\
tNodeEmbed~\cite{ijcai2019-640} & \cmark & \xmark & \cmark  & \cmark & \cmark \\
DySAT~\cite{sankar2020dysat} & \cmark & \xmark &  \cmark & \xmark  & \cmark \\
\midrule
\textbf{our framework} & \cmark & \cmark &  \cmark & \cmark  & \cmark \\

\bottomrule
\end{tabularx}
}
\end{table}

\section{Related Work} \label{sec:related-work}
\subsubsection*{Snapshot-based approaches.} The majority of temporal embedding approaches break down the graph into graph-time series based on the application time-scale (1 month, etc.) up to a certain timestamp $t$, and then derive features from them to make inference on graphs at $t+1$. 
One direction is to look into the most recent snapshot, for instance, DANE~\cite{li2017attributed} proposes to embed both nodes and the associated attributes in the graph by minimizing the loss of reconstruction of the snapshot at a given timestamp $t$: $\frac{1}{2}\Sigma_{i,j}\mathbf{A}^{(t)}_{ij}||y_i-y_j||^2$, and update the embeddings for snapshot at $t+1$ based on the change of graph structure and node attributes.
DynGEM~\cite{goyal2018dyngem} adopts the deep auto-encoder to generate the nonlinear embeddings from the snapshot at $t-1$ while addressing stability.
TIMERS~\cite{zhang2018timers} models the relative changes in adjacency matrices between snapshots and leverages incremental SVD to derive embeddings.
A more popular direction is to track back a certain number of snapshots from time $t$. Primarily, these approaches first derive node embeddings from each individual tracked snapshot and then merge them through specific operation. There are also works that do both jointly.    
Dyngraph2vec~\cite{goyal2019dyngraph2vec} leverage totally $l$ snapshots to predict the snapshot at $t+1$. It leverages various deep architectures (\ie, auto-encoder, RNN) to derive latent features by minimizing loss of reconstruction error: $||f(\mathbf{A_t}, \mathbf{A}_{t+1}, \cdots, \mathbf{A}_{t+l}) - \mathbf{A}_{t+l+1}||^{2}_{F}$.
tNodeEmbed~\cite{ijcai2019-640} is an end-to-end framework based on node embeddings derived from individual snapshots using static methods. The embeddings are merged by minimizing the loss of specific tasks (\ie, link prediction and node classification) through LSTM.
DySAT~\cite{sankar2020dysat} leverages the notion of self-attention to compute node representations by jointly employing graph structural property and temporal dynamics.

\subsubsection*{Sequential-interaction-based approaches.}
There is another line of works that studies the sequential interaction between nodes in the graph.
CTDNE~\cite{CTDNE-WWW18} is the first approach to learn embeddings directly from the stream of timestamped edges at the finest temporal granularity.
In that work, they proposed the notion of temporal walks and used it for embeddings~\cite{CTDNE-WWW18}.
More recently, node2bits~\cite{jin2019node2bits} expanded on this idea by incorporating features in the temporal walks and hashing them. 
This was shown to better capture the notion of structural similarity that lies at the heart of role-based embeddings~\cite{roles2015-tkde}.
Alternatively, some other work has modeled the node-specific temporal dynamics as the point process where the probability of interaction is represented through different intensity functions.
HTNE~\cite{zuo2018embedding} proposes to model the node evolution through the Hawkes process.
JODIE\cite{kumar2019predicting} models the sequential interaction in bipartite graphs to predict the change of embedding trajectory over time instead of interaction probability. 
CTDNE, HTNE and JODIE are designed to handle continuously sequential data, which is not the scope of this paper.

\section{Preliminaries}
\label{sec:prelim}
We summarize symbols and notations used in this work in Table~\ref{table:symbols}. Some important notions are given as follows.

\begin{table}[t!]
\centering
\vspace{-0.2cm}
\caption{Summary of notation}
\vspace{-3mm}
\centering 
\fontsize{7}{7.5}\selectfont
\setlength{\tabcolsep}{5pt} 
\label{table:symbols}
\vspace{-0.1cm}
\def\arraystretch{1.25} 
\begin{tabularx}{1.0\linewidth}{@{}rX@{}}
\toprule

$\mathcal{G} = \{G_1, G_2, \cdots \}$
& a graph time-series  \\
$G_t=(V_t, E_t)$ & a directed and weighted temporal network from $\mathcal{G}$ with $|V_t|$ nodes and $|E_t|$ temporal edges \\
$\mA_t$ & 
adjacency matrix for graph $G_t$ at time $t$ \\

$G_R=(V,E_R)$ & the weighted temporal reachability graph \\
$N^R_i$ & the set of nodes that are temporally reachable fro node $i$ \\

$\tau$ & window size representing the timespan of edges\\
$\epsilon$ & window size representing the number of edges\\
$\alpha$ & the decay factor in the temporal summary graph model\\
$\theta$ & the decay factor in the temporal embedding smoothing\\

$f$ & arbitrary base embedding method \\

$d$ & dimensionality of the embedding \\
$\mathbf{Z}$ & $|V|\times d$ embedding matrix\\

\bottomrule
\end{tabularx}
\vspace{-0.35cm}
\end{table}

\begin{Definition}[Temporal Graph]
Let $V$ be a set of vertices, and $E \subseteq V \times V \times \RR^{+}$ be the set of temporal edges between vertices in $V$. Each edge $(u,v,t)$ has a unique time $t \in \RR^{+}$.
\end{Definition}

When edges represent contacts---a phone call, an email, or physical proximity---between two entities at a specific point in time, we have a particular type of evolving network structure~\cite{ferreira2002models,bhadra2003complexity}.  
A temporal walk in such a network represents a sequence of contacts that obeys time.  
That is, if each edge represents a contact between two entities, then a path represents a feasible route for a piece of information.
\begin{Definition}[Temporal Walks]
A temporal walk from $u$ to $w$ in $G=(V,E)$ is a sequence of edges $e_1, \ldots, e_k$ such that $e_1\!=\!(u_1,u_2,t_1), \ldots, e_k\!=\!(u_k, u_{k+1}, t_k)$ where $t_j < t_{j+1}$ for all $j = 1$ to $k$.
We say that $u$ is \textit{temporally connected} to $w$ if there exists such a temporal walk.
\end{Definition}
This definition echoes the standard definition of a path, but adds the additional constraint that paths must respect time, i.e., follow the directionality of time.
Temporal walks are inherently asymmetric because of the directionality of time.
The notion of temporal walks has been recently used in embedding methods~\cite{CTDNE-WWW18}.

\section{Framework}
\label{sec:framework}
The framework in this paper provides a fundamental basis for studying different temporal network representations and the utility of these for generalizing embedding methods to temporal network data.
In particular, the framework serves as a basis to adapt and generalize existing static embedding methods for temporal networks by incorporating temporal dependencies into the learning component used by the different embedding methods.

\begin{figure*}[ht!]
\centering
\includegraphics[width=0.98\textwidth]{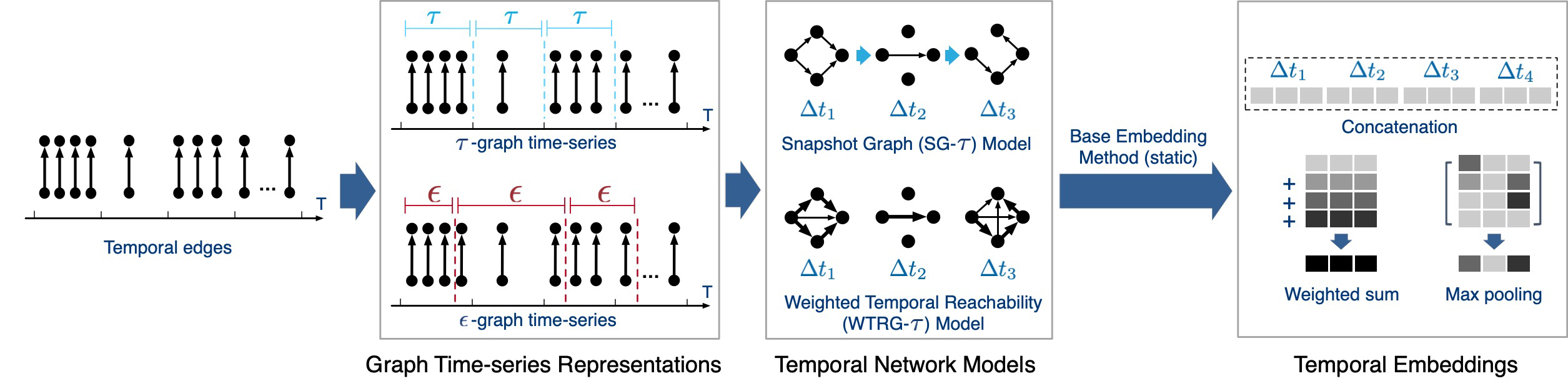}
\vspace{-0.2cm}
\caption{
Framework Overview. 
In the first component of the framework (Sec.~\ref{sec:graph-based-rel-time-series}), we derive a time-series of graphs from the stream of timestamped edges using either an application-specific time-scale $\tau$ (\eg, 1 day) or by using a fixed number of edges $\epsilon$ for each graph in the time-series.
Next, given the $\{\tau,\epsilon\}$-graph time-series representation, we use a temporal network model from Sec.~\ref{sec:temporal-network-models} to incorporate the temporal dependencies and temporal weights.
Finally, we use an arbitrary base embedding method to learn a time-series of embeddings and then leverage a temporal fusion mechanism to obtain the final temporal embeddings (Sec.~\ref{sec:temporal-embeddings}).
As we will show in Sec.~\ref{sec:exp-framework-variants-vs-state-of-the-art}, this general framework achieves comparable and often better predictive performance than state-of-the-art dynamic embedding methods.
}
\label{fig:overview}
\vspace{-.25cm}
\end{figure*}

As shown in Figure~\ref{fig:overview}, the proposed temporal network representation framework has the following main components.
First, given the continuous stream of timestamped edges $e_1,e_2,e_3,\ldots,e_{t-1},e_t,\ldots$, we derive the time-series of graphs 
(Section~\ref{sec:graph-based-rel-time-series}).
Second, given the graph-based time-series, we use one of the temporal network models to incorporate the temporal dependencies (Section~\ref{sec:temporal-network-models}).
Third, we use the framework to generalize existing embedding methods, effectively enabling the new dynamic variants of these methods to learn more accurate and appropriate
time-dependent embeddings.
(Section~\ref{sec:temporal-embeddings}).
Finally, in Section~\ref{sec:embedding-time-series-models} 
we also briefly describe a few general methods that can be used  over the resulting time-dependent embeddings for improving temporal prediction tasks.

\subsection{Graph Time-Series Representations}
\label{sec:graph-based-rel-time-series}
We formally introduce two approaches for deriving a time-series of graphs from the stream of timestamped edges.

\subsubsection{$\tau$-graph time-series}
The $\tau$-graph time-series representation is used by the vast majority of previous work~\cite{hisano2016semisupervised, kamra2017deepdualnetwork}.

\begin{Definition}[$\tau$-graph time-series]
\label{def:graph-time-series-time-based}
Given the continuous stream of timestamped edges $e_1,e_2,e_3,\ldots,e_{t-1},e_t,\ldots$ 
and let $\tau$ denote the application time scale representing a unit of time (such as a minute, day, week, etc.),
we define a $\tau$-graph time-series $\mathcal{G}^{\tau} = \{G_1,\ldots,G_k,\ldots,G_t\}$ such that $G_1$ consists of all edges within the first time scale (period) $\tau$, 
$G_2$ consists of all edges within the next time period $\tau$, and so on. 
Thus, each graph contains edges within a specific period of time $\tau$.
More formally, let $t_0$ denote the timestamp of the first edge in the temporal network (stream of timestamped edges) and $\tau$ is the application time-scale (\eg, 1 month), then 
\begin{equation} \label{eq:graph-time-series-time-based-application-time-scale} 
E_k = \big\{ (i,j,t) \in E \;|\; t_0+k\tau > t \geq t_0+(k-1)\tau \big\}
\end{equation}
Temporal models that use a time-based graph time-series are denoted with the suffix ``-$\tau$''.
\end{Definition}
The \emph{time-based} is used in Definition~\ref{def:graph-time-series-time-based} to denote that each graph represents a specific period of time (\eg, 5 minutes, 1 hour, 1 day).

\subsubsection{$\epsilon$-graph time-series}
\label{sec:omega-graph-time-series-N-edges}
While most work uses the previous approach for deriving the graph time-series, we introduce a new alternative based on the idea of using a fixed number of edges.
In particular, we propose a new approach that derives a time-series of graphs $\mathcal{G}^{\epsilon} = \{ G_1,\ldots,G_k,\ldots,G_t \}$ such that each $G_k$ consists of $\epsilon$ edges (Definition~\ref{def:graph-time-series-edge-based}) and therefore $|E_k|=\epsilon, \forall k$.
More formally,
\begin{Definition}[$\epsilon$-graph time-series]
\label{def:graph-time-series-edge-based}
Given the continuous stream of timestamped edges $e_1,e_2,e_3,\ldots,e_{t-1},e_t,\ldots$ 
and let $\epsilon$ denote a fixed number of temporal edges in the stream (ordered by time),
we define a graph time-series $\mathcal{G}^{\epsilon} = \{G_1,\ldots,G_k,\ldots,G_t\}$ such that $|E_k|=\epsilon$, for all $k=1,2,\ldots$. 
Hence, $G_1$ consists of the first $\epsilon$ edges $E_1 =\{e_1,e_2,\ldots,e_{\epsilon}\}$
whereas $G_2$ consists of the next $\epsilon$ edges $E_2=\{e_{\epsilon+1},\ldots,e_{2\epsilon}\}$, and so on.
More formally, $E_t$ is defined as follows:
\begin{align} \label{eq:omega-graph-time-series-edge-set}
E_t &= \bigcup_{i=(t-1)\epsilon+1}^{t\epsilon} e_i \;=\; \big\{e_{(t-1)\epsilon+1},\ldots,e_{t\epsilon}\big\}
\end{align}
Temporal models that use the notion of an $\epsilon$-graph time-series are denoted with the suffix ``$-\epsilon$''.
\end{Definition}\noindent
Note in both cases $E_1 \cup \cdots \cup E_t = E$.
To the best of our knowledge, this paper is the first to introduce this notion of an $\epsilon$-graph time-series with a fixed number of edges.
This approach has several important advantages:
\begin{itemize}[leftmargin=*]

\item The $\epsilon$-graph time-series representation enables us to control for the confounding factor of edge frequency since the number of edges is fixed for all $t$.
Therefore, the $\epsilon$-graph time-series enables us to focus on the structural properties over time since if $|E_t|=\epsilon$ for all $t$, then embeddings learned from any $E_t$ are expected to be relatively similar since they are all learned from the same number of edges, and therefore have the same opportunity to give rise to similar structure.
This is in contrast to the $\tau$-graph time-series representation, where we are unable to determine if a difference in embeddings (from $t$ to $t+1$)
is due to the number of edges,
or more importantly, due to changes in the structure over time.

\item Understand the network and structural properties better since the same number of edges is used in each snapshot graph in the time-series.

\item Avoid issues with learning when the number of edges that occur over time significantly varies with respect to time and therefore 
is highly dependent on the time-scale.
For instance, in many real-world data, the amount of edges that occur during a given month (or week, etc) is significantly different than another month.
\end{itemize}

\subsection{Temporal Network Models} \label{sec:temporal-network-models}
Now we introduce temporal network models that incorporate the temporal dependencies into the graph time-series representations to learn more effective time-dependent embeddings.

\subsubsection{Snapshot Graph (SG) Model}
This model simply leverages the $\{\tau,\epsilon\}$-graph time-series representation directly without encoding any additional temporal information into the representation.
Hence, the temporal information (edge timestamps) associated with the edges in any graph $G_t$ are effectively ignored/discarded. 
In other words, the sequential interactions (edges) between nodes in each graph $G_t$ in the time-series are ignored.
This model incorporates the temporal dependencies at the level of the graph, that is, we know that $G_{t-1}$ occurred before $G_{t}$ and so on.
The snapshot model that leverages the $\tau$-graph time-series is denoted as SG-$\tau$ whereas the snapshot model that uses $\epsilon$-graph time-series is denoted as SG-$\epsilon$.

\subsubsection{Temporal Summary Graph (TSG) Model} 
The temporally summary graph model incorporates the temporal dependencies by deriving a weighted summary graph from the graph-based time series $\mathcal{G}$~\cite{rossi2010drc-modeling,rossi2012drc} where the more recent edges are assigned larger weights than those in the distant past. 
More formally, let $\mA_1,\mA_2,...,\mA_t,...,\mA_T$ be a time-series of adjacency matrices of the graph time-series constructed using either Definition~\ref{def:graph-time-series-time-based} or Definition~\ref{def:graph-time-series-edge-based}.
Furthermore, let $\mA_t(i,j)$ denote the $(i,j)$ entry of $\mA_t$.
We define the general \emph{weighted temporal summary graph} (TSG) model as
\begin{equation} \label{eq:temporal-summary-graph}
\textstyle    \mS = \sum_{t=1}^{T} f(\mA_t, \alpha)
\end{equation}\noindent
where $f$ is a decay function for temporally weighting the edges (nonzeros), $\alpha$ is the decay factor ranging in $(0,1)$, $T$ is the total number of graphs in the time-series, and $\mS$ is the weighted temporal summary graph.
In this work, we define $f$ as an exponential decay function since this was previously shown to perform well~\cite{rossi2012drc}, then we obtain
\begin{equation} \label{eq:temporal-summary-graph-exp-func}
\textstyle \mS = \sum_{t=1}^{T} (1-\alpha)^{T-t} \mA_t
\end{equation}\noindent
Furthermore, the weight for an edge $(i,j)$ is simply $\mS(i,j) = \sum_{t=1}^{T} (1-\alpha)^{T-t} \mA_t(i,j)$.
Hence,
\begin{equation}
\mS = (1-\alpha)^{T-1}\mA_1 + (1-\alpha)^{T-2}\mA_2 + \cdots + (1-\alpha)\mA_{t-1} + \mA_t
\end{equation}

Notice that the weighted TGS model can leverage the previous time-scale based $\tau$-graph time-series representation or the new $\epsilon$-graph time-series representation that uses a fixed number of edges to represent each temporal graph in the time-series.

\subsubsection{Time-Series of Weighted Summary Graphs}\label{sec:weighted-TSG-graph-time-series}
We can also derive another temporal model based on using the weighed TSG temporal model to derive a time-series of weighted temporal summary graphs based on a temporal lag $\ell$.
Let $\ell$ denote the time-series lag and let $T$ denote the number of graphs in a time-series using either $\{\tau,\epsilon\}$-graph time-series representation, then
\begin{equation} \label{eq:time-series-weighted-temporal-summary-graphs}
\mS_t = \sum_{k=t-\ell}^{t} f(\mA_k, \alpha),\quad \forall t=\ell+1,\ldots,T
\end{equation}\noindent
Hence, Eq.~\ref{eq:time-series-weighted-temporal-summary-graphs} gives us the following \emph{temporally weighted summary graph} time-series:
\begin{equation}
\mS_t, \mS_{t+1}, \ldots, \mS_{T}  
\end{equation}\noindent
Alternatively, instead of using all available graphs in the initial time-series, we can use only the $L$ most recent graphs. 
For example, suppose $\mathcal{G}^{\epsilon}=\{G_t\}^{T}_{t=1} = \{G_1,\ldots,G_T\}$ is an $\epsilon$-graph time-series with $T$ graphs.
Instead of using all $T$ graphs, we can leverage only the most recent $L$ graphs, hence,
\begin{equation} \label{eq:most}
\mathcal{G}^{\epsilon}=\{G_t\}^{T}_{t=T-L+1} = \{G_{T-L+1}, \ldots,G_T\}
\end{equation}\noindent
The idea of leveraging only the most recent graphs in the time-series was first explored in~\cite{rossi2012drc} and can be leveraged for any of the proposed temporal models in this section.

\subsubsection{Temporal Reachability Graph (TRG) Model}
The temporal reachability graph (TRG) is a graph derived from the timestamped edge stream where a link is added between two nodes if they are temporally connected. 
More formally, an edge $(u, v)$ in the TRG model indicates the existence of a temporal walk from $u$ to $v$ in the original graph.
The formal definition is given as follows.

\begin{Definition}[Temporal Reachability Graph] 
Given an interval $\mathcal{I} \in \mathbb{R}^+$, the temporal reachability graph $G_R=(V, E_R)$ is defined as a directed graph where the edge $(u, v)\in E_R$ denotes the existence of a temporal walk leaving $u$ and arriving $v$ within that interval. We denote the number of edges in $\mathcal{I}$ as $\omega$ (which could be defined based on $\{\tau, \epsilon\}$-graph time-series).
\label{def:trg}
\end{Definition}

\begin{figure}[h!]
\vspace{-2mm}
\captionsetup[subfigure]{justification=centering}
\centering
\begin{subfigure}[t]{0.15\textwidth}
\centering
\includegraphics[width=0.98\textwidth]{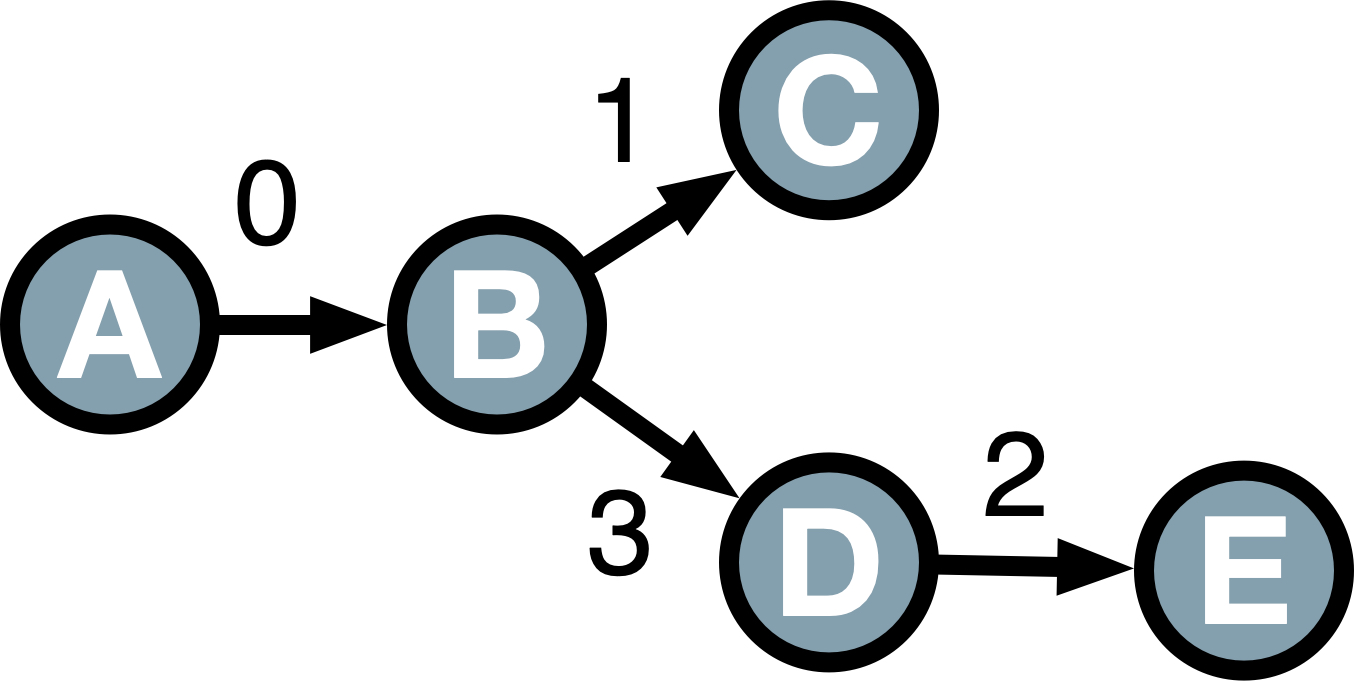}
\caption{A temporal graph}
	\label{fig:trg_example_a}
\end{subfigure}
~
\begin{subfigure}[t]{0.15\textwidth}
\centering
\includegraphics[width=0.98\textwidth]{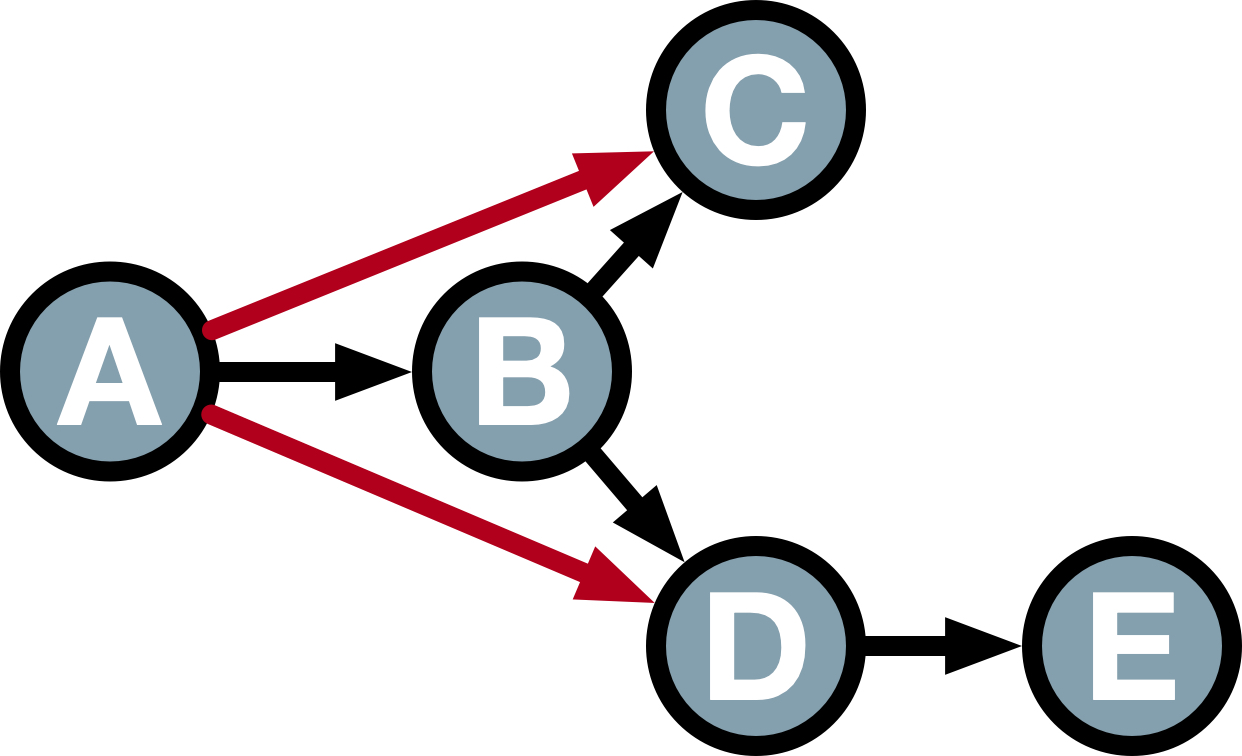}
\caption{TRG}
\label{fig:trg_example_b}
\end{subfigure}
~
\begin{subfigure}[t]{0.15\textwidth}
\centering
\includegraphics[width=0.98\textwidth]{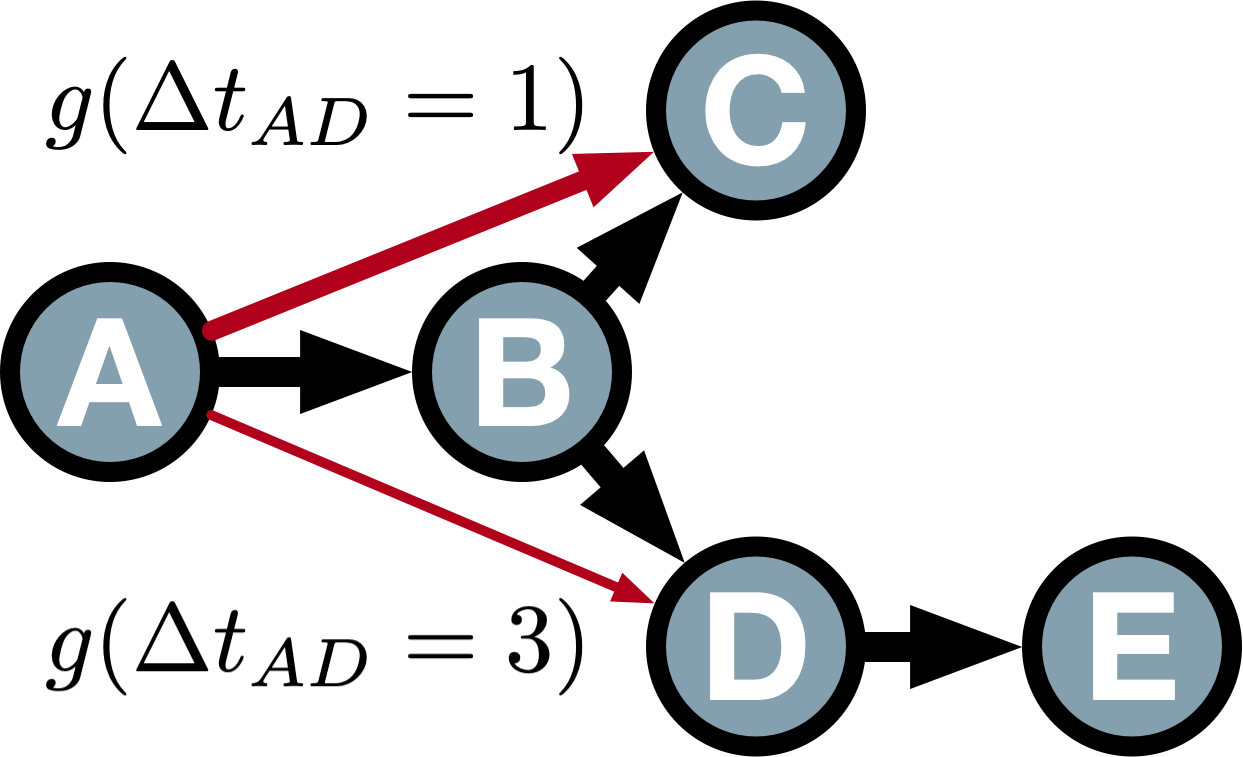}
\caption{Weighed TRG}
\label{fig:trg_example_c}
\end{subfigure}
\vspace{-0.3cm}
\caption{A temporal graph (a) and its temporal reachability modeling TRG (b) and WTRG (c). (b) An edge in the vanilla TRG represents a temporally-valid walk. The red edges represents the length-2 walks $\{A,B,C\}$ and $\{A,B,D\}$ in the original graph (c) WTRG extends TRG by assigning weights to indicate the temporal closeness \eg, $\{A,B,C\}$ has higher weights than $\{A,B,D\}$ as $C$ is temporally closer to $A$ than $D$.
}
\label{fig:trg_example}
\vspace{-.15cm}
\end{figure}

A TRG is a static unweighted graph where each edge indicates a temporally-valid walk reaching from the source to the destination. However, it does not capture the strength of reachability. 
For example in Fig.~\ref{fig:trg_example_a}, the walk $\{A, B, C\}$ takes two timestamps while $\{A, B, D\}$ takes four. Intuitively $D$ is harder to reach than $C$ from node $A$. However, in the derived TRG shown in~\ref{fig:trg_example_b}, all the edges are equally important with weight $1$, which makes the strength indistinguishable. This would potentially affect the proximity-based embedding methods as they are based on the closeness of nodes in the graph.

To overcome this drawback, we propose an extension of TRG called Weighted TRG (WTRG) that encapsulates the strength of reachability in the graph weights. We define the strength of reachability between a pair of nodes $(i, j)$ as a function of both the number of temporally-valid paths and the timestamp difference. Specifically, the weighting function is given as follows.
\begin{equation}
\textstyle g_{i, j} = \sum_{w\in\mathcal{W}}e^{-(\Delta t_{i,j}|w)}
\label{eq:trg_weight}
\end{equation}
where $w$ is a specific temporally-valid walk from $i$ to $j$, and $\Delta t_{i,j}$ denotes the temporal delay reaching from $i$ to $j$ along that walk. 
We depict the process of deriving WTRG in Algorithm~\ref{alg:temporal-reachability-graph-weight}.

\begin{algorithm}[t!]
\caption{Weighted Temporal Reachability Graph}
\small
\begin{algorithmic}[1]
\Procedure {TemporalReach}{$G = (V,E)$}

\State  Set $E_R = \emptyset$
\State Sort $E_T$ in reverse time order
\While{next edge $(i,j,t) \in E$} 

\For{$(k,t_k) \in N^{R}_j$} 
\State $E_R\leftarrow E_R \cup \{(i, k)\}$ 
\State $g_{i,k} = g_{i,k} + e^{-(t_k-t)}$
\State $N_i^R \leftarrow N_i^R \cup \{(k,t_k)\}$
\EndFor

\State $E_R \leftarrow E_R \cup \{(i,j)\}$ 

\State $g_{i,j} = g_{i,j} + 1$ \Comment{$\Delta t_{i,j}=0$ as $i,j$ are adjacent}
\State $N_i^R \leftarrow N_i^R \cup \{(j,t)\}$

\EndWhile

\State \textbf{return} $G_R = (V,E_R,g)$

\EndProcedure
\end{algorithmic}
\label{alg:temporal-reachability-graph-weight}
\end{algorithm}

The cornerstone of the algorithm is the temporally-reachable neighborhood $N_i^R$ that records nodes that can be reached by $i$ as well as the latest timestamps associated with the temporal paths. Formally, we define $N_i^R$ as follows.

\begin{Definition}[Temporally reachable neighborhood]
Given a node $i$, its temporally reachable neighborhood $N^R_i$ is defined as the set of tuples $\{(j, t_j)\}$ where $j$ is the node reachable from $i$ following a temporally-valid walk and $t_j$ is the timestamp of the edge reaching $j$ in that walk.
\end{Definition}

Given an input temporal edge $(i,j,t)$, Algorithm~\ref{alg:temporal-reachability-graph-weight} loops through reachable neighbors in $N_i^R$ to add edges in $E_R$ and updates the weights based on Eq.~\eqref{eq:trg_weight}(line 5-8). It also adds $(i,j)$ to the WTRG as well as the immediate weight (line 9-11). Overall, the computational complexity of the algorithm is $\mathcal{O}(|E|\max d(N^R))$, where $\max d(N^R)$ is the maximum degree of a node in WTRG.

While the derived WTRG can be dense with huge amounts of reachable neighbors, we show that this number is bounded by $\omega$, which is the size of the interval associated with the WTRG (Def.~\ref{def:trg}). Accordingly, the computational complexity of the algorithm can be denoted as $\mathcal{O}(|E|\omega)$.

\begin{property}\label{prop:walk-temporal-reach-graph}
The number of edges in $G_R$ is bounded by the number of temporally-valid walks in $G$.
\end{property}
Based on Def.~\ref{def:trg}, an edge $(u,v)\in E_R$ indicates a temporally-valid walk reaching from $u$ to $v$ in $G$. However, this edge could correspond to multiple unique temporal walks with different intermediate nodes and associated timestamps, therefore, $|E_R|$ is no more than the number of temporally-valid walks in $G$.

\begin{Claim}
Let $N^R_i$ denote the temporally reachable nodes of $i$, $\Delta(G_R) = \max\{d(N^R_1),\ldots,d(N^R_n)\}$ is the maximum degree of a node in $G_R$, and $\omega$ is the window size.
Then
\begin{equation}
|N^R_i| \leq \Delta(G_R) \leq \omega
\end{equation}
\end{Claim}
According to Def.~\ref{def:trg}, a TRG is comprised by edges within the interval with size $\omega$. These edges comprise upto $\omega$ different temporal walks originating from a specific node $i$. Therefore, based on Property~\ref{prop:walk-temporal-reach-graph}, the number of edges originating from node $i$ is bounded by the number of temporally-valid walks, which is $\omega$.

Given the graph time-series representations, we create two variants of weighted temporal reachability graph models, which are \methodweakT and \methodweakN that leverages the $\tau$-graph time-series and $\epsilon$-graph time-series, respectively.
To combine the embeddings over the graph time-series, we follow Algorithm~\ref{alg:temporal-reachability-embedding-framework}.
\begin{algorithm}[t!]
\small
\begin{algorithmic}[1]
\caption{General Framework for Temporal Embeddings
}
\label{alg:temporal-reachability-embedding-framework}
\Ensure 
$\epsilon$ or $\tau$ for deriving the graph time-series representation, 
base embedding method $f$ (\eg, GraphWave, role2vec)
\medskip
\State Construct a graph time-series $\mathcal{G} = \{G_1, G_2, \ldots, G_t\}$ using a graph time-series representation $\{\tau,\epsilon\}$ from Section~\ref{sec:graph-based-rel-time-series}.
\medskip

\State Initialize $\mZ_0$ to all zeros
\smallskip
\For{{\bf each} $G_t \in \mathcal{G}$} \Comment{for $t=1,2,\ldots$}
\multiline{Use Alg.~\protect\ref{alg:temporal-reachability-graph-weight} to derive the temporal reachability graph for $G_t$
\medskip
}

\multiline{Compute node embedding matrix $\mZ_t$ using the base embedding method $f$ with the temporal reachability graph from Alg.~\protect\ref{alg:temporal-reachability-graph-weight}
\medskip
}

\multiline{Concatenate or aggregate (using sum, mean, etc.) the embedding matrix, \eg, 
$\bar{\mZ}_{t} = (1-\theta)\bar{\mZ}_{t-1} + \theta \mZ_{t}$ 
where $\bar{\mZ}_t$ is the temporally weighted embedding using the above exponential weighting kernel $\mathbb{K}(\cdot)$ and $0 \leq \theta \leq 1$ is a hyperparameter controlling the importance of past information relative to more recent (Section~\ref{sec:embedding-time-series-models}).
}
\EndFor
\multiline{\textrm{\bf return} $\bar{\mZ}_{t}$ (temporally weighted embeddings using $\mathbb{K}$ and $\theta$) \emph{or} $\mZ = \big[\;\mZ_1 \; \mZ_2 \,\cdots\, \mZ_t\;\big]$ (concatenated embeddings)
}

\end{algorithmic}
\end{algorithm}
\vspace{-0.2cm}
\subsection{Temporal Embeddings}
\label{sec:temporal-embeddings}
\subsubsection{Base embedding methods}
\label{sec:base-embedding}
Given the graph time-series representation and temporal model from the previous components (Section~\ref{sec:graph-based-rel-time-series}-\ref{sec:temporal-network-models}), the proposed approach can leverage any existing static embedding method to derive time-dependent node embeddings that capture the important temporal dependencies between the nodes as well as the temporal structural (role-based)~\cite{dbmm2012mining} and proximity-based properties~\cite{from-comm-to-structural-role-embeddings}.
Hence, the framework can always leverage the state-of-the-art embedding method to learn time-dependent node embeddings.

We use the proposed framework to generalize a wide variety of static base embedding methods including both community-based and role-based structural node embedding methods~\cite{from-comm-to-structural-role-embeddings}. 
Namely, they are:
{\bf (1)} LINE~\cite{line},
{\bf (2)} Node2vec~\cite{node2vec},
{\bf (3)}~Graph2Gaussian~\cite{bojchevski2018deep},
{\bf (4)}~struc2vec~\cite{struc2vec},
{\bf (5)}~Role2vec~\cite{role2vec},
{\bf (6)}~Graphwave~\cite{donnat2018learning}, and
{\bf (7)}~multilens~\cite{jin2019latent}. 
We provide the detailed configuration of each individual method in the appendix for reproducibility (Appendix~\ref{sec:appendix-embedding-configuration}).
Among these static methods, (1-3) are community/proximity-based and (4-6) are role-based. 
(7) is a hybrid that is based on structural similarity of node-central subgraphs. 

\subsubsection{Temporal fusion}\label{sec:embedding-time-series-models}
Given the time-series of node embeddings $\{\mZ_t\}_{t=1}^{T}$, how do we use them for prediction and other downstream applications?

\smallskip\noindent\textbf{Concatenation of the time-series of node embeddings:} 
Given a time-series of embeddings, one simple approach to obtain a final embedding for prediction is to simply concatenate the embeddings as follows:
$\mZ = \big[ \mZ_1 \cdots \mZ_T \big]$.
However, we can also weight the embeddings based on time.
Alternatively, we can moderate the influence of the embeddings by devoting a larger embedding size to the more recent embeddings (or obtaining new low-rank approximation of the embeddings that occur in the distant past, this would effectively compress the more distant ones further since they are not as important as the more recent ones, which we allow a larger embedding dimension). This is another way to bias the embeddings toward more recent events for temporal prediction tasks.

\smallskip\noindent\textbf{Temporally weighting the node embeddings:}
\label{sec:embedding-time-series-models-temporally-weighted-model}
Concatenate or aggregate (using sum, mean, etc.) the embedding matrix, \eg, 
$\bar{\mZ}_{t} = (1-\theta)\bar{\mZ}_{t-1} + \theta \mZ_{t}$ 
where $\bar{\mZ}_t$ is the temporally weighted embedding using the above exponential weighting kernel $\mathbb{K}(\cdot)$ and $0 \leq \theta \leq 1$ is a hyperparameter controlling the importance of past information relative to more recent.

\section{Experiments}
\label{sec:exp}
In this section, we systematically investigate the different graph time-series representations (Section~\ref{sec:exp-T-vs-N}, temporal network models (Section~\ref{sec:exp-temporal-model-comparison}), and the new dynamic node embedding methods generalized using the proposed framework (Section~\ref{sec:exp-framework-variants-vs-state-of-the-art}).

Experiments are carefully designed to investigate the effectiveness of the proposed framework for generalizing existing static embedding methods to learn time-dependent embeddings.
We also investigate the effectiveness of the proposed methods introduced in each of the components of the framework.
More specifically, we investigate the following questions:
\begin{itemize}
\item \textbf{Q1} Does the $\tau$-graph time-series representation that previous methods use perform better than the proposed $\epsilon$-graph time-series representation?

\item \textbf{Q2} What temporal models are most useful for incorporating temporal dependencies into static embedding methods? Does one temporal model consistently perform better than others?
Is there a clear ranking of temporal models?

\item \textbf{Q3} 
Are the new dynamic node embedding methods
(generalized via the framework)
from the framework useful for temporal prediction? 
How do they compare to the state-of-the-art dynamic embedding methods?
\end{itemize}

\subsection{Experimental Setup}

\textbf{Temporal Network Data} In these experiments, we use a variety of real-world temporal networks from SNAP~\cite{snapnets} and NR~\cite{nr}.
The statistics and properties are summarized in Table.~\ref{table:data_stats}. We provide detailed data description in Appendix~\ref{sec:appendix-data-description} and data preprocessing in Appendix~\ref{sec:appendix-data-preprocessing}.

\begin{table}[t!]
\vspace{-0.3cm}
\centering
\caption{Network statistics and properties}
\label{table:data_stats}
\vspace{-0.3cm}
\centering 
{\small
\setlength{\tabcolsep}{6pt} 
\begin{tabular}{lrr C{1.2cm} L{1.5cm}  }
\toprule
\textbf{Data} & $|V|$ & $|E|$ & Type & Timespan \\ \hline
enron & 151 & 50,572 & Unipartite & 38 months \\
bitcoin & 3,783 & 24,186 & Unipartite & 63 months \\
wiki-elec & 7118  & 107,071 & Unipartite & 47 months \\
stackoverflow & 24,818  & 506,550 & Tripartite & 79 months \\
fb-forum & 899  & 33,720 & Unipartite & 24 weeks \\
reallity-call & 6,809  & 52,050 & Unipartite & 16 weeks \\
wiki-edit & 8,227 & 157,474 & Bipartite & 32 days \\
contact-dublin & 10,972  & 415,912 & Unipartite & 69 days \\
\bottomrule
\end{tabular}
}
\vspace{-0.2cm}
\end{table}

\smallskip\noindent\textbf{Configuration} 
We consider the task of link prediction over time and systematically compare the performance of different temporal network models and representations.
Given a set of timestamped edges up to time $t$, the temporal link prediction task is to predict the future links that will form at time $t+1$.
We first construct a graph time-series representation $\mathcal{G} = \{G_{1}, G_{2}, \cdots, G_{k}, \cdots G_{t}\}$ based
on either an application-specific time-scale $\tau$ (\ie, $G_{k}\in \mathcal{G}^{\tau}$ represents the edges that occur within a temporal unit shown in Table~\ref{table:data_stats})
or an $\epsilon$-graph time-series representation where $G_{k}\in \mathcal{G}^{\epsilon}$ consists of the most recent $\epsilon$ edges.
For each $\{\epsilon, \tau\}$-graph time-series representation, we derive a temporal network model based on it, namely, SG, TSG, and WTRG.

For fair comparison, we ensure that embeddings derived from both $\epsilon$ and $\tau$-based temporal network models are used to predict links in the same hold-out test set at time $t+1$.
Specifically, we first follow the conventional setup and fix the time-scale of the $\tau$-based models to determine the testing links $E_{t+1}$ for prediction, which ensures each graph $G_k \in \mathcal{G}^{\tau}$ and $G_{t+1}$ are consistent with respect to the $\tau$ representation.
Then, we set $\epsilon=|E_{t+1}|$ for the $\epsilon$-based temporal models so that the graphs in the $\epsilon$-graph time-series $\mathcal{G}^{\epsilon} = \{G_1, G_2,\ldots,G_t\}$ and $G_{t+1}$ are also consistent with respect to the $\epsilon$ representation, where $|E_1| = |E_2| = \cdots = |E_{t+1}|$.
This ensures that each graph $G_k \in \mathcal{G}$ in the $\epsilon$-graph time-series $\mathcal{G} = \{G_1, G_2,\ldots,G_t\}$ consists of $\epsilon$ edges.

Using an arbitrary embedding method from the framework, we learn node embeddings and give them as input to a logistic regression model for prediction with regularization strength $1.0$ and stopping criteria $10^{-4}$.
Following~\cite{rossi2018deep}, we derive an embedding of an edge between node $i$ and $j$ by concatenating the node embeddings $\vz_i$ and $\vz_j$ to obtain an edge embedding defined as $\vz_{ij}=\big[\, \vz_i \; \vz_j \,\big]$.
For the time-series of node embeddings, we use the temporally weighted node embedding model from Section~\ref{sec:embedding-time-series-models-temporally-weighted-model} with $\theta=0.8$.
The TSG decay parameter $\alpha$ is set to $0.8$ for consistency. 

\begin{table}[h!]
\vspace{-0mm}
\centering
\setlength{\tabcolsep}{9.5pt}
\renewcommand{\arraystretch}{1.10} 
\caption{Overall ranking of the temporal network models (1st column).
Temporal models are ordered from best to worst.
Overall score is simply the sum of the \# of times each temporal model performed best overall (ranked first) across all evaluation criterion, base embedding methods, and graph datasets.
We observe that all the top-3 temporal network models are based on the new $\epsilon$-graph time-series representation that uses a fixed amount of edges (Section~\ref{sec:omega-graph-time-series-N-edges})
as opposed to the application time-scale ($\tau$) model that existing methods have used.
}
\vspace{-2.2mm}
\label{table:temporal-network-model-overall-ranking}
{
\small
\begin{tabularx}{1.0\linewidth}{@{} r ccc c}
\toprule
\vspace{-1mm}
& \multicolumn{3}{c}{\textsc{\# of First Ranks}} & \textsc{Overall} \\
\textsc{Temporal Model} & \textsc{AUC} & \textsc{ACC} & \textsc{F1} & \textsc{Score} \\
\midrule
\textbf{TSG-$\epsilon$} & 12 & 14 & 14 & \textbf{40} \\
\textbf{WTRG-$\epsilon$} & 14 & 9 & 9 & \textbf{32} \\
\textbf{SG-$\epsilon$} & 10 & 9 & 8 & \textbf{27} \\
\textbf{WTRG-$\tau$} & 8 & 9 & 9 & \textbf{26} \\
\textbf{SG-$\tau$} & 9 & 8 & 9 & \textbf{26} \\
\textbf{Static} & 0 & 5 & 5 & \textbf{10} \\
\textbf{TSG-$\tau$} & 3 & 2 & 2 & \textbf{7} \\
\bottomrule

\vspace{-1mm}
\textsc{Time-series} & \multicolumn{3}{c}{\textsc{\# of First Ranks}} & \textsc{Overall} \\
\textsc{Representation} & \textsc{AUC} & \textsc{ACC} & \textsc{F1} & \textsc{Score} \\
\midrule
\textbf{$\epsilon$-based models} & 36 & 32 & 31 & \textbf{99} \\
\textbf{$\tau$-based models} & 20 & 19 & 20 & \textbf{59} \\
\bottomrule

\end{tabularx}
}
\vspace{-1mm}
\end{table}

\subsection{Fixed number of edges ($\epsilon$) vs. time-scale ($\tau$)}
\label{sec:exp-T-vs-N}
In this section, we investigate the effectiveness of different graph time-series representations (\textbf{Q1}).
For each $\{\epsilon,\tau\}$-graph time-series representation, we select a temporal network model  $\{SG, TSG, WTRG\}$ and a base embedding method using the framework.
Therefore, we have $\{SG\text{-}\epsilon, TSG\text{-}\epsilon,WTRG\text{-}\epsilon\}$ dynamic variants using the $\epsilon$-graph time-series representation and $\{SG\text{-}\tau,TSG\text{-}\tau,WTRG\text{-}\tau\}$ using the $\tau$-graph time-series representation.

We evaluate the performance using AUC, ACC and F1 score. Let $\vy_{jk} \in \RR^{|\mathcal{M}|}$ denote the vector of AUC (or ACC, F1) scores of the temporal models $\mathcal{M}$ for an embedding method $f_j \in \mathcal{F}$ and graph dataset $k$.
Further, let $\pi(\vy_{jk}, M_i)$ denote the rank of the temporal model $M_i \in \mathcal{M}$ for a given embedding method $f_j$ and graph dataset $d_k \in \mathcal{D}$.
We define the score $s_i$ of temporal model $M_i$ as 
\begin{equation} \label{eq:total-scores-first-rank}
s_i = \sum_{d_k \in \mathcal{D}} \sum_{f_j \in \mathcal{F}} \mathbb{I}\big\{\pi(\vy_{jk}, M_i) = 1\big\}
\end{equation}\noindent
where $\mathbb{I}\{\pi(\vy_{jk}, M_i) = 1\}$ returns $1$ if $\pi(\vy_{jk}, M_i) = 1$ and $0$ otherwise.
In other words, $\pi(\vy_{jk}, M_i) = 1$ if the temporal model $M_i$ performs best for the given graph dataset $d_k$ and base embedding method $f_j$.
Therefore, $s_i$ is the total score of model $M_i$ based on the number of times the temporal model $M_i$ appeared first in the ranking across all base embedding methods and graph data sets.
We also compute an overall score over all evaluation criterion (AUC, ACC, and F1 score) by simply summing over each $s_i$ for all evaluation criterion. 
This follows from~\cite{zhang2007ml,sml-2018} and provides an intuitive ranking of the temporal network models based on the number of times each temporal model performed best.

The results are provided in Table~\ref{table:temporal-network-model-overall-ranking}.
Strikingly, we find that the top-3 temporal models are those that use the proposed $\epsilon$-graph time-series representation as opposed to the $\tau$-graph time-series representation used in previous work (which is based on an application specific time-scale).
This finding indicates that future work should instead use the proposed $\epsilon$-graph time-series to represent the temporal network.

\begin{Result}\label{res:omega-performs-better}
The proposed $\epsilon$-graph time-series representation based on a fixed number of edges $\epsilon$ significantly outperforms the time-scale based representation $\tau$ used by most existing methods as shown in Table~\ref{table:temporal-network-model-overall-ranking}.
\end{Result}

\begin{figure}[t!]
\centering
\includegraphics[width=0.49\linewidth]{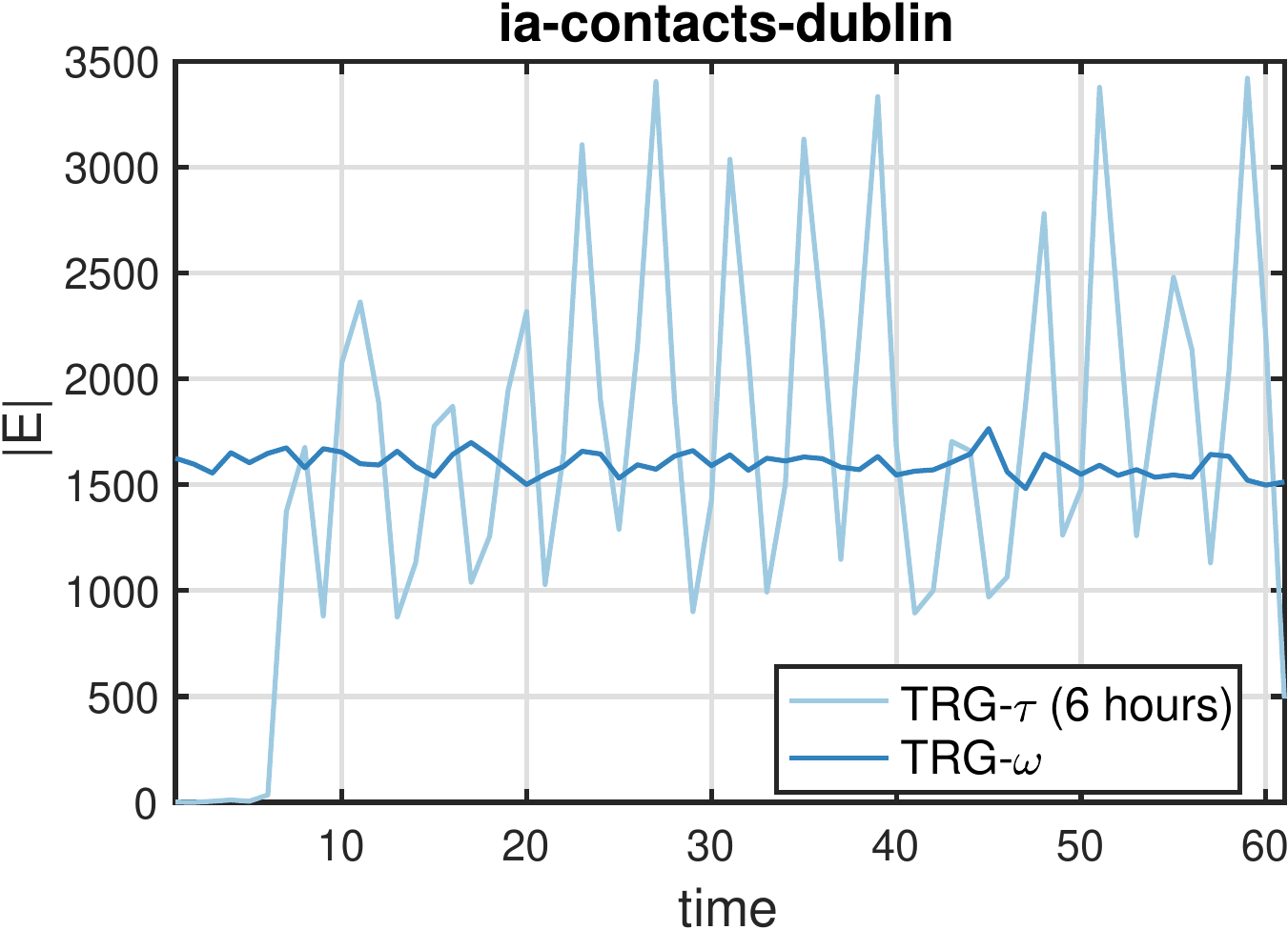}\hfill
\includegraphics[width=0.475\linewidth]{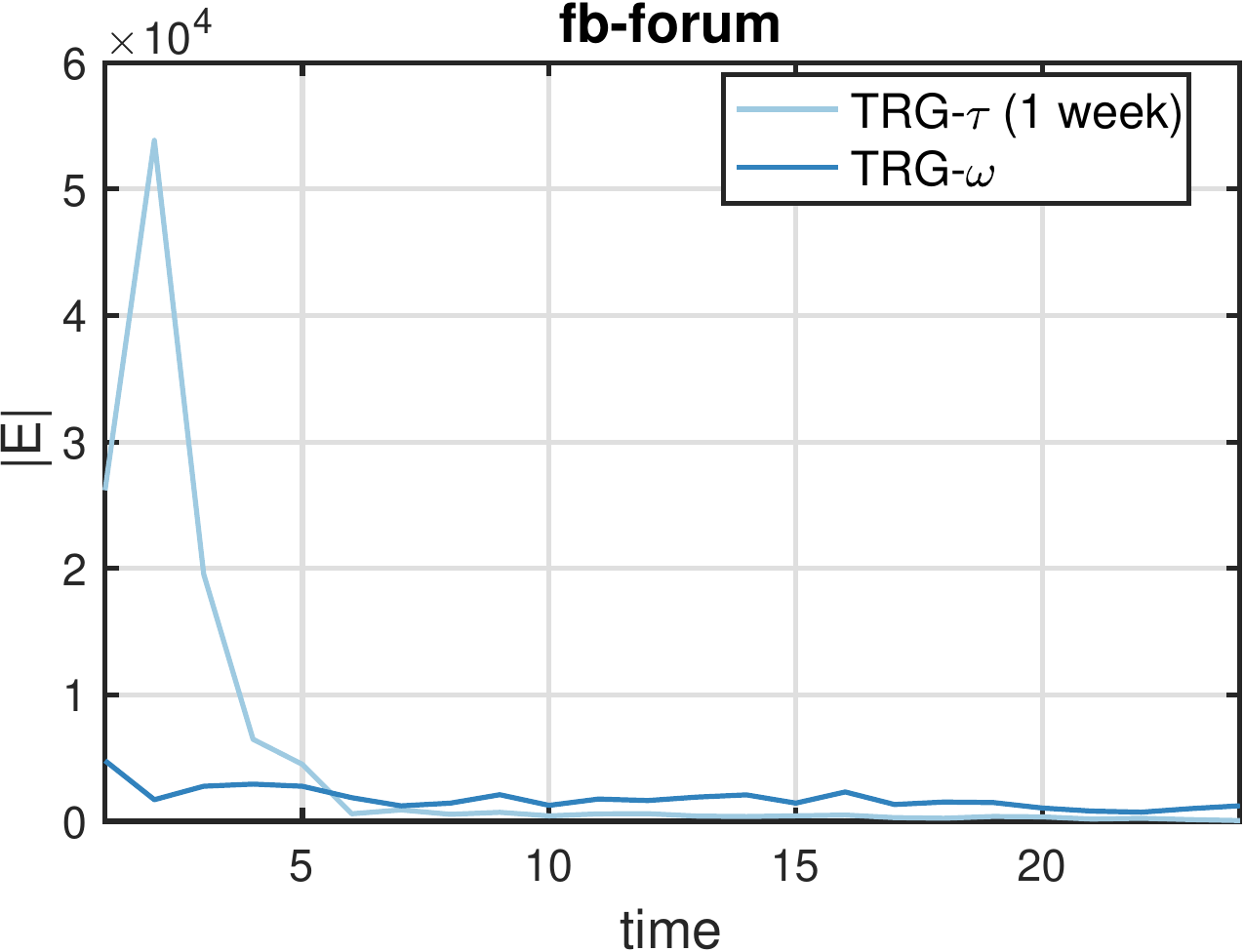}\hfill
\vspace{-2mm}
\caption{Comparing $\{\epsilon,\tau\}$-graph time-series representations combined with the WTRG-$\{\tau,\epsilon\}$
temporal models.
}
\label{fig:properties-comparing-graph-time-series-exp}
\vspace{-1mm}
\end{figure}

In Figure~\ref{fig:properties-comparing-graph-time-series} and~\ref{fig:properties-comparing-graph-time-series-exp}, we see a clear advantage of using the proposed $\epsilon$-graph time-series with a fixed $\epsilon$ number of edges compared to the $\tau$-graph time-series that is based on a specific application time-scale (\eg, 10 minutes, 1 day).
In particular, while the SG-$\epsilon$ model has a fixed number of edges over time as shown in Figure~\ref{fig:properties-comparing-graph-time-series}, the number of edges in SG-$\tau$ can significantly deviate with large spikes, even between consecutive graphs in the time-series.
This holds for the WTRG models as well as shown in Figure~\ref{fig:properties-comparing-graph-time-series-exp} where \methodweakN has a relatively stationary number of edges over time whereas \methodweakT is significantly impacted by large spikes in the number of edges over time with significantly more fluctuations.
The $\epsilon$-graph time-series representation generally benefits all temporal models as shown in Table~\ref{table:temporal-network-model-overall-ranking}.

\begin{Result} 
The structural properties from the $\epsilon$-graph time-series with a fixed $\epsilon$ number of edges are significantly more stable and robust compared to the $\tau$-graph time-series based on a specific application time-scale (\eg, 1 hour, 1 day).
\end{Result}

\subsection{Temporal Model Comparison} \label{sec:exp-temporal-model-comparison}
To answer \textbf{Q2}, we follow the formalization in \S~\ref{sec:exp-T-vs-N} to quantitatively evaluate and rank the temporal models according to their effectiveness with respect to prediction.\footnote{As an aside, we observed that the WTRG model outperforms the vanilla TRG model and use it throughout our experiments (see Appendix~\ref{sec:appendix-exp-wtrg} for more details).}
We show the performance of temporal network models with respect to individual datasets in Table~\ref{table:graphs-by-temporal-models-num-first-ranks} along with the overall ranking of them in Table~\ref{table:temporal-network-model-overall-ranking}.

Overall, the WTRG and SG model tend to perform well across all datasets. 
Moreover, the temporal models that are combined with the proposed $\epsilon$-graph time-series representation always outperform the other models, which is consistent with our previous findings from Section~\ref{sec:exp-T-vs-N}.
Notably, the TSG-$\epsilon$ model performs the best  while the TSG-$\tau$ performs the worst.
In addition, WTRG-$\epsilon$ performs the second best and is a close second to TSG-$\epsilon$.
This is due to the fact that the TSG and WTRG model encapsulates the temporal information into the graph time-series.
In TSG, larger edge weights represent the temporal strength of connection combined with the recency of timestamped edges.
Furthermore, the fluctuation of temporal edges in the $\tau$-graph time-series has a larger impact on the performance of the dynamic embedding methods that use these models.
On the other hand, the $\epsilon$-graph time-series representation does not have this issue and we observe that the TSG-$\epsilon$ model achieves the best performance overall. 
This indicates that with a proper graph time-series representation, the model is still capable to derive node embeddings that reflect both the temporal structure and the temporal recency and importance.
This result further confirms the importance of the $\epsilon$-graph time series representation.

\begin{table}[t!]
\vspace{-0.0mm}
\centering
\setlength{\tabcolsep}{3.5pt}
\caption{
Results comparing the temporal models across the different temporal network data.
Each $(i,j)$ is the \# of times temporal model $M_j \in \mathcal{M}$ in graph $G_i$ performed best (relative to the other models) across all base embedding methods $f \in \mathcal{F}$ and evaluation criterion.
For each dynamic graph, we bold the temporal model that performed best overall.
}
\label{table:graphs-by-temporal-models-num-first-ranks}
\vspace{-3.0mm}
{\footnotesize
\begin{tabularx}{\linewidth}{@{} r  c cccc cc H@{}}
\toprule
&  \textbf{TSG-$\epsilon$} &  \textbf{WTRG-$\epsilon$} &  \textbf{SG-$\epsilon$} &  \textbf{SG-$\tau$} &  \textbf{WTRG-$\tau$} &  \textbf{Static} &  \textbf{TSG-$\tau$} & \\ 
\midrule
\textsf{bitcoin} & \textbf{6} & \textbf{6} & 4 & 5 & 0 & 0 & 0 & \\
\textsf{stackoverflow} & 1 & 4 & 3 & 3 & \textbf{9} & 0 & 1 & \\
\textsf{enron} & 4 & 1 & 1 & 3 & \textbf{8} & 4 & 0 & \\
\textsf{wiki-elec} & 2 & 6 & \textbf{7} & 6 & 0 & 0 & 0 & \\
\textsf{fb-forum} & \textbf{10} & \textbf{10} & 0 & 1 & 0 & 0 & 0 & \\
\textsf{wikipedia} & \textbf{7} & 3 & 2 & 3 & 2 & 2 & 2 & \\
\textsf{reality-call} & 1 & 0 & 2 & 4 & \textbf{6} & 4 & 4 & \\
\textsf{contacts-dublin} & \textbf{9} & 2 & 8 & 1 & 1 & 0 & 0 & \\
\midrule
\textbf{overall score} & 40 & 32 & 27 & 26 & 26 & 10 & 7 & \\
\bottomrule
\end{tabularx}
}
\vspace{-2mm}
\end{table}

\begin{Result}
The temporal network model that performs the best is TSG-$\epsilon$ followed closely by WTRG-$\epsilon$ (Table~\ref{table:temporal-network-model-overall-ranking}).
\end{Result}
\noindent
Furthermore, WTRG and SG tend to also perform well across all base embedding methods and evaluation criterion. 
In addition, these models always perform best when combined with the $\epsilon$-graph time series.
We also notice that the proximity-based embedding methods~\cite{from-comm-to-structural-role-embeddings} tend to perform badly when using the \methodweakN model. 
From Table~\ref{table:graphs-by-temporal-models-num-first-ranks}, we observe that the temporal models that leverage the $\epsilon$-graph time series representation perform well across all datasets. 
However, these temporal models perform especially well on datasets such as \texttt{fb-forum} and \texttt{contacts-dublin}, where significant spikes and fluctuation are observed (Figure~\ref{fig:properties-comparing-graph-time-series} and Figure~\ref{fig:properties-comparing-graph-time-series-exp}).

\subsection{Dynamic Embeddings: Framework Variants vs. State-of-the-art}
\label{sec:exp-framework-variants-vs-state-of-the-art}
To answer \textbf{Q3}, we first use the framework to derive new dynamic embedding methods (by selecting the representation, temporal model, base embedding method, and so on from the framework, which uniquely defines a new dynamic embedding method), then we compare the performance of the resulting dynamic embedding methods from the framework to the state-of-the-art dynamic embedding methods on all 8 datasets.
One would of course expect that the state-of-the-art methods for dynamic node embeddings will outperform the dynamic embedding methods generalized by our framework.
This is because the state-of-the-art methods are typically more complex and have been designed specifically for learning such dynamic node embeddings.
For these experiments, we use seven state-of-the-art dynamic embedding methods as baselines, including CTDNE~\cite{CTDNE-WWW18}, node2bits~\cite{jin2019node2bits}, DANE~\cite{li2017attributed}, DynGem~\cite{goyal2018dyngem} TIMERS~\cite{zhang2018timers}, DynAE/DynAERNN~\cite{goyal2019dyngraph2vec}, and DySAT~\cite{sankar2020dysat}.
For reproducibility, we provide detailed configuration in Appendix~\ref{sec:appendix-dyn-method-configuration}.

\begin{figure}[h!]
\centering
\hspace{-2mm}\includegraphics[width=0.7\linewidth]{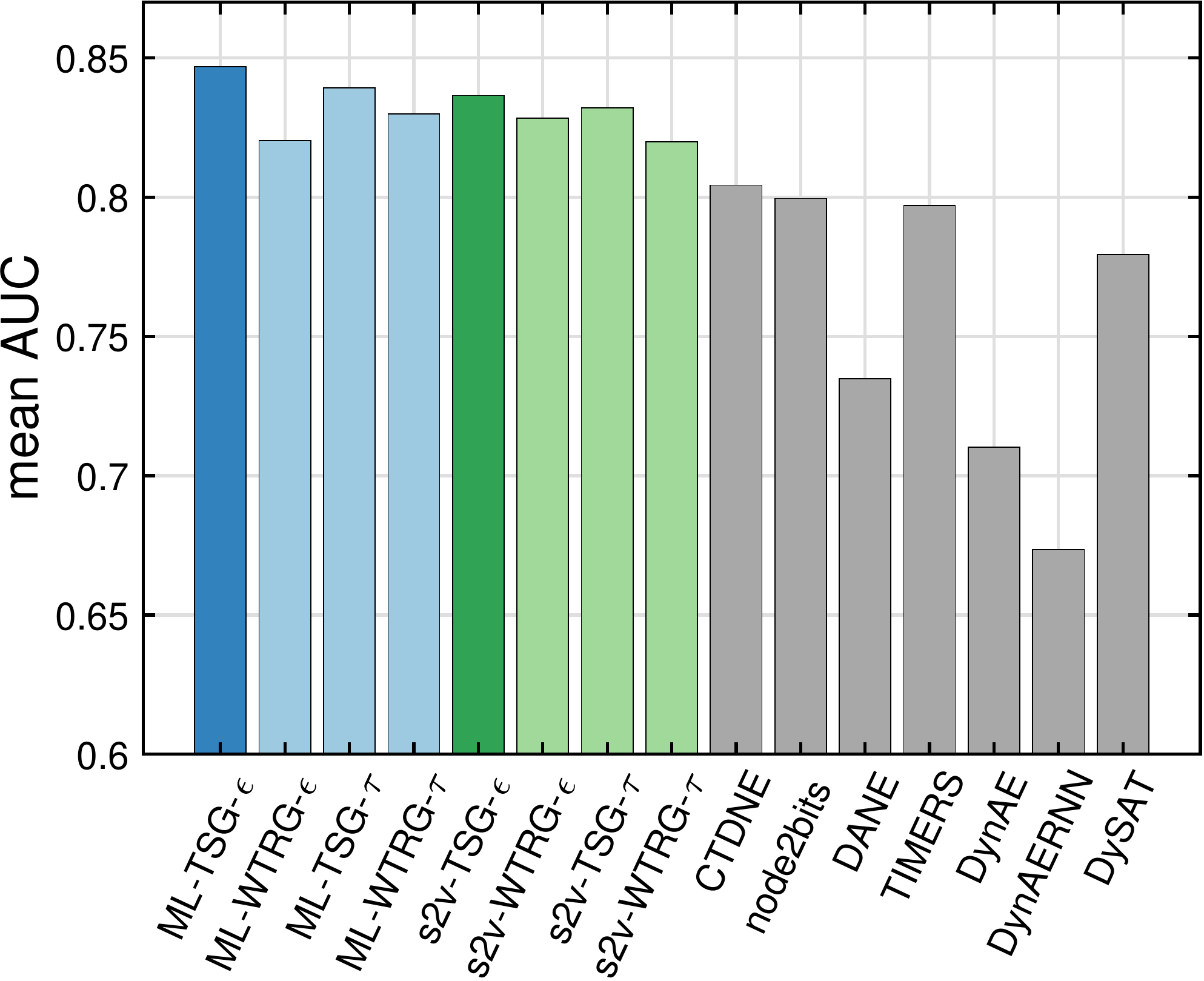}

\vspace{-2mm}
\caption{Results comparing the state-of-the-art dynamic embedding methods to the methods generalized by the framework (ML $=$ multilens, s2v $=$ struc2vec).
}
\label{fig:exp-dynamic-embeddings-state-of-the-art-vs-framework}
\vspace{-.05cm}
\end{figure}

In Figure~\ref{fig:exp-dynamic-embeddings-state-of-the-art-vs-framework}, we show the mean AUC for each method where the average is taken over all graphs investigated.
As representative dynamic embedding methods from the proposed framework, we use four dynamic embedding variants of struc2vec 
(s2v-TSG-$\epsilon$, s2v-WTRG-$\epsilon$, s2v-WTRG-$\epsilon$) 
and two other variants of MultiLENS (ML-TSG-$\epsilon$, ML-WTRG-$\epsilon$).
Strikingly, we observe that the dynamic embedding methods from the framework
outperform the state-of-the-art methods that are designed particularly for temporal graphs and time-series prediction.

\begin{Result}
The dynamic embedding methods from the proposed framework (Sec.~\ref{sec:framework}) perform better than the more complex state-of-the-art dynamic embedding methods as shown in  Fig.~\ref{fig:exp-dynamic-embeddings-state-of-the-art-vs-framework} and Table~\ref{table:mean-gain-dynamic-embeddings-framework-vs-state-of-the-art}.
\end{Result}

In Table~\ref{table:mean-gain-dynamic-embeddings-framework-vs-state-of-the-art}, we report the mean gain in AUC for each of the dynamic embedding methods from the framework compared to each of the seven state-of-the-art dynamic embedding methods.
Strikingly, in all cases, the dynamic embedding methods from the framework outperform the seven state-of-the-art dynamic embedding methods.
In particular, ML-TSG-$\epsilon$ performs best with a mean gain of 12.34\% followed by s2v-TSG-$\epsilon$
with a gain in AUC of 10.97\%.

Notably, we can use the proposed framework as a blackbox to generalize any static embedding method to a more powerful and predictive dynamic embedding method.
More strikingly, the framework is simple and powerful enough to leverage any base embedding method, yet these new methods still outperform the state-of-the-art dynamic embedding methods on most datasets, and in most cases, without complicated hyperparameter tuning or time-consuming learning stages.
In other words, these dynamic embedding methods from the framework achieve better predictive performance than existing state-of-the-art dynamic node embedding methods that are significantly more complex and developed specifically for such temporal prediction tasks.
These results demonstrate the utility of the proposed framework and motivates its use in future research for developing better dynamic node embedding methods as well as evaluating the utility of more sophisticated and complex methods.

\begin{table}[t!]
\vspace{-0.0mm}
\centering
\setlength{\tabcolsep}{1.5pt}
\caption{
Mean gain of the dynamic embedding methods from the framework relative to each of the more complex state-of-the-art methods.
Last column is the mean gain over the seven state-of-the-art methods.
}
\label{table:mean-gain-dynamic-embeddings-framework-vs-state-of-the-art}
\vspace{-2.5mm}
{
\footnotesize
\begin{tabularx}{\linewidth}{@{}l ccccccc r H@{}}
\toprule
&  \textbf{CTDNE} &  \textbf{n2b} &  \textbf{DANE} &  \textbf{TIMERS} &  \textbf{DynAE} &  
\textbf{DynAR} & 
\textbf{DySAT} & 
\textsc{Gain} \\ 
\midrule
\textbf{ML-TSG-$\epsilon$} & 5.30\% & 5.93\% & 15.26\% & 6.26\% & 19.25\% & 25.75\% & 8.66\% & \textbf{12.34\%} & \\
\textbf{ML-WTRG-$\epsilon$} & 1.99\% & 2.60\% & 11.64\% & 2.92\% & 15.50\% & 21.80\% & 5.24\% & 8.81\% & \\
\textbf{ML-TSG-$\tau$} & 4.34\% & 4.97\% & 14.21\% & 5.30\% & 18.17\% & 24.61\% & 7.67\% & 11.32\% & \\
\textbf{ML-WTRG-$\tau$} & 3.17\% & 3.79\% & 12.93\% & 4.11\% & 16.84\% & 23.21\% & 6.46\% & 10.07\% & \\
\midrule
\textbf{s2v-TSG-$\epsilon$} & 4.01\% & 4.63\% & 13.84\% & 4.96\% & 17.79\% & 24.21\% & 7.33\% & \textbf{10.97\%} & \\
\textbf{s2v-WTRG-$\epsilon$} & 2.98\% & 3.60\% & 12.72\% & 3.93\% & 16.63\% & 22.98\% & 6.27\% & 9.87\% & \\
\textbf{s2v-TSG-$\tau$} & 3.44\% & 4.06\% & 13.22\% & 4.39\% & 17.15\% & 23.53\% & 6.74\% & 10.36\% & \\
\textbf{s2v-WTRG-$\tau$} & 1.92\% & 2.54\% & 11.56\% & 2.86\% & 15.43\% & 21.72\% & 5.18\% & 8.74\% & \\
\bottomrule
\multicolumn{9}{l}{\TT
\footnotesize
$^{*}$ Note ML=MultiLENS, s2v=struc2vec, n2b=node2bits, DynAR=DynAERNN}\\
\end{tabularx}
}
\vspace{-1mm}
\end{table}

\section{Conclusion}
\label{sec:conclusion}
This work proposed a general and flexible framework that can serve as a basis for generalizing existing or future state-of-the-art static embedding methods, as well as studying different graph-based time-series representations, temporal network models, and base embedding methods.
Despite the recent increasing interest in temporal networks in the field of representation learning, there has been relatively little work that systematically studies the properties of temporal network models and the graph time-series representations that lie at their heart.
This works attempts to fill this gap by proposing a powerful framework that can be used to naturally generalize any existing or future state-of-the-art static embedding approach to a family of fully dynamic embedding methods.
Specifically, we propose the $\epsilon$-graph time-series representation that uses a fixed number of edges as opposed to the traditional way of deriving a graph time-series based on a---sometimes arbitrary---time-scale (e.g., 1 day or 1 week). 
Most importantly, the $\epsilon$-graph time-series representation is useful for applications where it is important to model and capture the \emph{structural changes} of the graphs over time whereas the $\tau$-graph time-series is better for capturing edge \emph{frequency changes} (as opposed to structural changes).
We find that the generalized dynamic embedding methods that leverage the proposed $\epsilon$-graph time-series outperform those that use the conventional $\tau$-graph time-series.
Furthermore, our proposed framework gives rise to new dynamic embedding methods by combining the $\{\epsilon,\tau\}$-graph time-series representations, new temporal models, and base static embedding methods.  
We find that the generalized embedding methods from the framework that leverage the proposed $\epsilon$-graph time-series representation along with the proposed WTRG and TSG models perform the best across nearly all datasets.
We show that these dynamic embedding methods from our framework outperform recent state-of-the-art dynamic embedding methods that are more complex.
Finally, we expect the findings of this work will be useful in understanding and developing better embedding methods for temporal networks.

\balance
\bibliographystyle{ACM-Reference-Format}
\bibliography{paper-v1}

\newpage
\appendix
{\center \textbf{\LARGE Supplementary Material on Reproducibility}}

\section{Data description}
\label{sec:appendix-data-description}
The detailed description of the experimental graph datasets is given as follows.

\begin{itemize}
\item \texttt{enron}\footnote{\label{nr}\url{http://networkrepository.com}} records email exchanging between employees of Enron from May, 1999 to June, 2002.

\item \texttt{bitcoin}\footnote{\label{snap}\url{https://snap.stanford.edu/data/}} is a who-trusts-whom network of people who trade using bitcoins from Nov, 2010 to Feb., 2017. We study the user connectivity by dropping the edge signs.

\item \texttt{wiki-elec}\footref{nr} contains the voting history based on the Wikipedia page edit history from March, 2004 to Jan., 2008.

\item \texttt{stackoverflow}\footref{snap} is a temporal network consisting of three types of interactions on the stack exchange web site Math Overflow: a user answers questions, a user comments on questions, and a user comments on answers.

\item \texttt{wiki-edit}\footnote{\url{https://github.com/srijankr/jodie}} is a public bipartite dataset containing one month of edits made by users in the Wikipedia page.

\item \texttt{fb-forum}\footref{nr} is the Facebook-like Forum network that records users' activity in the forum.

\item \texttt{contacts-dublin}\footref{nr} is a human contact network where nodes represent humans and edges between them represent proximity (i.e., contacts in the physical world).

\item \texttt{reality-call}\footref{nr} is a subgraph of the reality mining study where nodes are participants and edges are phone calls. 
\end{itemize}

\section{Data preprocessing}
\label{sec:appendix-data-preprocessing}
We learn node embeddings from the graph time-series starting from roughly $\frac{1}{3}$ of the timespans. For example, for the \texttt{bitcoin} dataset, we train the classifier based on node embeddings derived from month $20$ to month $25$ out of 63 months, inclusive. This ensures that there are sufficient training edges to predict links in the following month. 
For all datasets, we perform training on the first $6$ graphs and predict links on the $7$th graph.
Depending on the time-scale shown in Table~\ref{table:data_stats}, they represent 6 months (\texttt{enron, bitcoin, wiki-elec and stackoverflow}), weeks (fb-forum and reality-call), or days (wiki-edit and contact-dublin).
We create evaluation examples from the links in the $7$th graph and
an equal number of randomly sampled pairs of unconnected nodes as negative samples~\cite{sankar2020dysat}.

\section{Base embedding method configuration}
\label{sec:appendix-embedding-configuration}
We configured all the baselines to achieve the best performance according to the respective papers. 
For all the baselines that are based on random walks (\ie, node2vec, struc2vec),
we set the number of walks to 20 and the maximum walk length to $L=20$.
For node2vec, we perform grid search over $p,q \in \{0.25, 0.50, 1, 2, 4\}$ as mentioned in~\cite{node2vec} and report the best performance. 
For LINE and Multi-Lens, we incorporate 2nd-order proximity in the graph.
For role2vec, we leverage the node degree as the feature for roles.
For Graphwave, we perform the method to automatically select the scaling parameter with exact heat kernel matrix calculation.
For all embedding approaches, we aim to generate final embedding with dimension $K=128$ for evaluation. Therefore, for concatenation fusion, the dimension of each individual graph time-series representation is $\frac{128}{T}$ where $T$ is the total number of graphs in the time-series. And for weighted summarization fusion, the individual dimensions are fixed to be $128$. 

\section{Dynamic embedding method configuration}
\label{sec:appendix-dyn-method-configuration}
For the state-of-the-art dynamic embedding methods, we follow the configuration given by the paper/code repository. For methods that are based on deep learning, we perform 5-fold cross validation with grid search to tune the hyperparameters for optimal performance. 
Specifically, for CTDNE, we set the number of walks to be $10$, the walking length to be $20$ for each node.
For node2bits, we set the method to perform short-term temporal random-walk with temporal scope to be $3$. The the number of walks and the walking length are set to be the same as CTDNE.
For DANE, we leverage both the offline computation model to derive node embeddings based on the first $6$ graphs, and the online model to derive node embeddings for the $6$th graph based on the first $5$. We set the intermediate embedding dimensions to be 100 for both models and report the best performance.
For TIMERS, we set the tolerance threshold value that is used to restart the optimal SVD calculation to be $0.17$ as provided in the code repository.
For DyAE/DyAERNN, we leverage the 2-layer auto-encoder/decoder with 400 and 200 units, respectively. We set the regularization hyperparameter to be $10^{-6}$, bounding ratio for number of units in consecutive layers to be $0.3$ as suggested in the paper, and perform grid search in the range of $\pm 10\%$ of the default value.
In the learning stage, the sgd learning rate is set to be $10^{-6}$ with minibatch size to be $100$.
Lastly, for DySAT, we leverage the base model with default hyperparameters provided in the code repository, and perform grid search in the range of $\pm 10\%$ of the default values.

\section{Impact of WTRG} \label{sec:appendix-exp-wtrg}
Next we study the effectiveness of WTRG model over the vanilla TRG model. 
As WTRG incorporates the strength of reachability in edge weights, we consider embedding methods that handles weighted graphs, namely, they are node2vec, struc2vec and multilens. We run these methods on two datasets using both TRG and WTRG with $\tau$-graph time series as shown in Table~\ref{table:exp_wtrg_trg}.

\begin{table}[!ht]
\centering
\caption{Performance of WTRG over TRG (both models use $\tau$-graph time series)}
\label{table:exp_wtrg_trg}
\vspace{-0.1cm}
\centering 
{\footnotesize
\setlength{\tabcolsep}{6pt} 
\def\arraystretch{1.1} 
\begin{tabular}{llrl|ll}
\toprule
& \multicolumn{3}{c}{\texttt{bitcoin} } & \multicolumn{2}{c}{\texttt{wiki-elec}} \\\hline
\multicolumn{1}{c}{Method} & Metric & TRG & WTRG & TRG  & WTRG    \\\hline
\multirow{3}{*}{node2vec} 
& AUC & 0.9214 & {\bf 0.9239} & {\bf 0.7348} & 0.7344 \\
& ACC & 0.8294 & {\bf 0.8412} & {\bf 0.6171} & 0.6144  \\
& F1  & 0.8285 & {\bf 0.8408} & {\bf 0.5909} & 0.5889 \\ \hline
\multirow{3}{*}{struc2vec} 
& AUC & 0.9274 & {\bf 0.9301} & 0.7840 & {\bf 0.7933} \\
& ACC & 0.7959 & {\bf 0.8109} & 0.6583 & {\bf 0.6703} \\
& F1  & 0.7925 & {\bf 0.8081} & 0.6388 & {\bf 0.6534} \\    \hline
\multirow{3}{*}{multilens} 
& AUC & 0.9226 & {\bf 0.9389} & 0.8106 & {\bf 0.8143}  \\
& ACC & 0.8656 & {\bf 0.8793} & 0.7438 & {\bf 0.7539}  \\
& F1  & 0.8655 & {\bf 0.8792} & 0.7385 & {\bf 0.7493}  \\  
\bottomrule
\end{tabular}
}
\end{table}

The first observation from Table~\ref{table:exp_wtrg_trg} is that structure-based embedding methods tend to outperform node2vec, the proximity-based method. In addition, we observe that WTRG improves most embedding methods in link prediction, except for node2vec on \texttt{wiki-elec} dataset. One possible reason is that the random walker in WTRG are more likely to visit nodes that are close in time, and thus limiting the derived embeddings to incorporate distant neighborhood information. We put this deep study of WTRG in the future work. Nevertheless, for embedding methods that are based on structural information, WTRG outperforms TRG by $0.8\%$ in AUC ,$1.3\%$ in ACC, and $1.4\%$ in F1 score on average.

\begin{Result}
\emph{Structural role-based} embedding methods generalized via WTRG typically perform better than proximity-based embedding methods.
\end{Result}

\label{sec:supplementary}

\end{document}